\documentclass[10pt,twocolumn,letterpaper]{article}
\pdfoutput=1

\usepackage{color}

\usepackage{url}

\definecolor{noah_color}{rgb}{0.2,.64,0}
\definecolor{kb_color}{rgb}{1,.14,0}
\definecolor{sean_color}{rgb}{1,.5,0}
\definecolor{paul_color}{rgb}{.6,.25,.9}
\definecolor{todo_color}{rgb}{.8,.8,.1}
\definecolor{new_color}{rgb}{.1,.7,.8}

\newenvironment{packed_item}{
\begin{itemize}
  \setlength{\itemsep}{1pt}
  \setlength{\parskip}{0pt}
  \setlength{\parsep}{0pt}
}{\end{itemize}}

\newcommand{\leaveout}[1]{}

\newcommand{\new}[1]{#1}

\newcommand{\etal}{{et~al.}}

\newcommand{\tablesize}[1]{{\footnotesize{#1}}}

\newcommand{\OSone}{\text{OpenSurfaces}\xspace}
\newcommand{\OStwo}{\textbf{Materials in Context Database}\xspace}

\newcommand{\OStwoshort}{\text{MINC}\xspace}
\newcommand{\minc}{\text{MINC}\xspace}

\newcommand{\segment}{\text{segment}\xspace}
\newcommand{\segments}{\text{segments}\xspace}

\newcommand{\alexnet}{\text{AlexNet}\xspace}

\newcommand{\decaf}{\text{DeCAF}\xspace}
\newcommand{\vgg}{\text{VGG-16}\xspace}

\newcommand{\googlenet}{\text{GoogLeNet}\xspace}

\newcommand{\krahenbuhl}{\text{Kr\"ahenb\"uhl}\xspace}
\newcommand{\website}{\url{http://minc.cs.cornell.edu/}\xspace}

\usepackage{amsopn}

\newcommand{\given}{\,|\,}

\usepackage{cvpr}
\usepackage{times}
\usepackage{microtype}
\usepackage{epsfig}
\usepackage{graphicx}
\usepackage{amsmath}
\usepackage{amssymb}
\usepackage{multirow}
\usepackage{booktabs}
\usepackage{nth}
\usepackage{array}
\usepackage{xcolor}

\usepackage[pagebackref=true,breaklinks=true,letterpaper=true,colorlinks,bookmarks=false]{hyperref}

\cvprfinalcopy

\def\cvprPaperID{1974} 

\begin{document}

\newcommand*\samethanks[1][\value{footnote}]{\footnotemark[#1]}

\title{Material Recognition in the Wild with the \OStwo}
\author{%
Sean Bell\thanks{Authors contributed equally}
\qquad
Paul Upchurch\samethanks
\qquad
Noah Snavely
\qquad
Kavita Bala
\\
Department of Computer Science, Cornell University
\\
{\tt\small \{sbell,paulu,snavely,kb\}@cs.cornell.edu}
}

\maketitle

\begin{abstract}

Recognizing materials in real-world images is a challenging task.  Real-world
materials have rich surface texture, geometry, lighting conditions, and
clutter, which combine to make the problem particularly difficult.  In this paper,
we introduce a new, large-scale, open dataset of materials in the wild, the
\OStwo (\OStwoshort), and combine this dataset with deep learning to achieve
material recognition and segmentation of images in the wild.

\OStwoshort is an order of magnitude larger than previous material databases,
while being more diverse and well-sampled across its 23 categories.
Using \OStwoshort, we train convolutional neural networks (CNNs) for two
tasks: classifying materials from patches, and simultaneous material
recognition and segmentation in full images.  For patch-based classification on
\minc we found that the best performing CNN architectures can achieve
85.2\% mean class accuracy.  We \new{convert} these trained CNN classifiers
into an efficient fully convolutional framework combined with a fully connected
conditional random field (CRF) to predict the material at every pixel in an
image, achieving 73.1\% mean class accuracy.
Our experiments demonstrate that having a large, well-sampled dataset such as
MINC is crucial for real-world material recognition and segmentation.


\end{abstract}


\vspace{-4pt}
\section{Introduction}

Material recognition plays a critical role in our understanding of and
interactions with the world. To tell whether a surface is easy to walk on, or
what kind of grip to use to pick up an object, we must recognize the materials
that make up our surroundings.
Automatic material recognition can be useful in a variety of applications,
including robotics, product search, and image editing for interior design. But
recognizing materials in real-world images is very challenging. Many categories
of materials, such as fabric or wood, are visually very rich and span a diverse
range of appearances. Materials can further vary in appearance due to lighting
and shape. Some categories, such as plastic and ceramic, are often smooth and
featureless, requiring reasoning about subtle cues or context to differentiate
between them.

\begin{figure}[tb]
\begin{center}
 \includegraphics[width=0.483\textwidth]{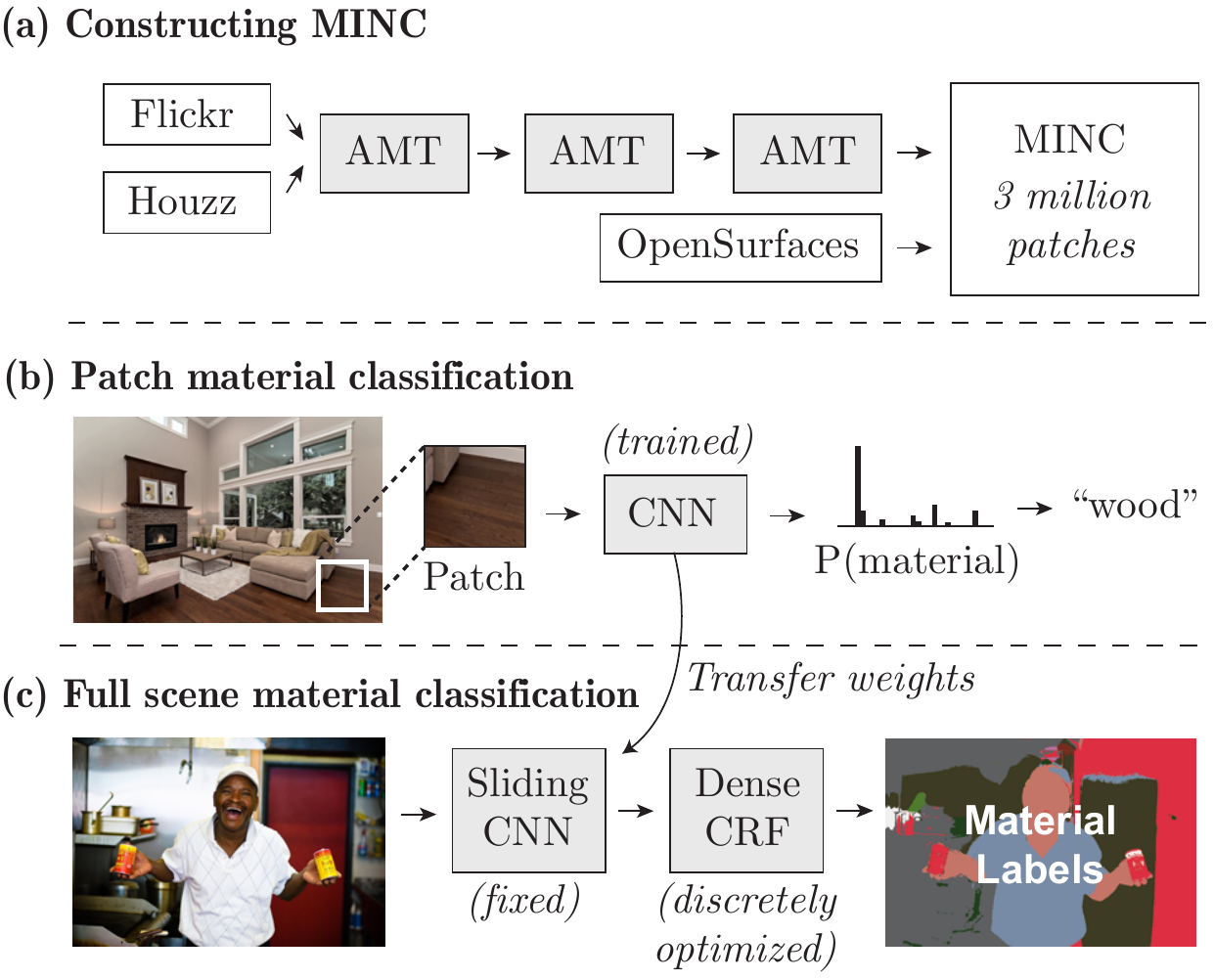}
\end{center}
  \vspace{-10pt}
   \caption{%
     \textbf{Overview.} \textbf{(a)} We construct a new dataset by combining
     OpenSurfaces~\cite{bell13opensurfaces} with a novel three-stage Amazon
     Mechanical Turk (AMT) pipeline.  \textbf{(b)} We train various CNNs on
     patches from MINC to predict material labels.  \textbf{(c)} We transfer the
     weights to a fully convolutional CNN to efficiently generate a probability
     map across the image; we then use a fully connected CRF to predict the
     material at every pixel.
   }
\label{fig:arch-overview}
\end{figure}


Large-scale datasets (e.g., ImageNet~\cite{russakovsky2014imagenet},
SUN~\cite{xiao2014sun,patterson2014sun} and Places~\cite{zhou2014places})
combined with convolutional neural networks (CNNs) have been key to recent
breakthroughs in object recognition and scene classification.  Material
recognition is similarly poised for advancement through large-scale data and
learning.  To date, progress in material recognition has been facilitated by
moderate-sized datasets like the Flickr Material Database
(FMD)~\cite{sharan2009material}. FMD
contains ten material categories, each with 100 samples
drawn from Flickr photos.
These images were carefully selected to illustrate a wide range of
appearances for these categories. FMD has been used in research on new
features and learning methods for material perception and
recognition~\cite{liu2010exploring,hu2011toward,qi2012pairwise,sharan-IJCV2013}.
While FMD was an important step towards material recognition, it is
not sufficient for classifying materials in real-world imagery.  This
is due to the relatively small set of categories, the relatively small
number of images per category, and also because the dataset has been
designed around hand-picked iconic images of materials.
%
The OpenSurfaces dataset~\cite{bell13opensurfaces} addresses some of these
problems by introducing \new{105,000} material segmentations from real-world
images, and is significantly \new{larger} than FMD.  However, in OpenSurfaces
many material categories are under-sampled, with only tens of images.

\begin{figure*}[t]
  \begin{center}
    \begin{tabular}{@{}c@{\hskip 1em}c@{\hskip 1em}c@{\hskip 1em}c@{\hskip 1em}c@{\hskip 1em}c@{\hskip 1em}c@{\hskip 1em}c@{}}
      \includegraphics[width=0.105\textwidth]{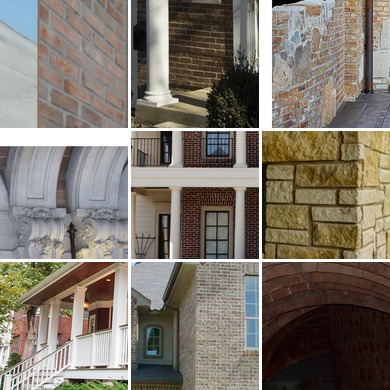} &
      \includegraphics[width=0.105\textwidth]{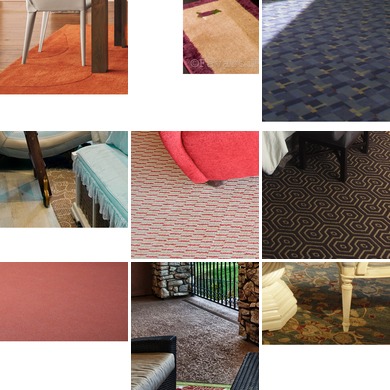} &
      \includegraphics[width=0.105\textwidth]{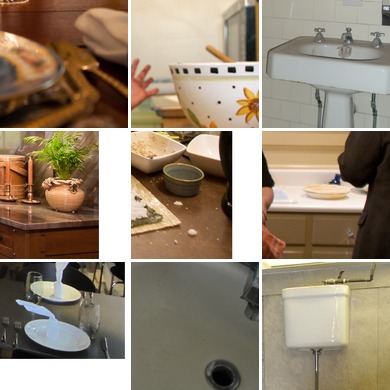} &
      \includegraphics[width=0.105\textwidth]{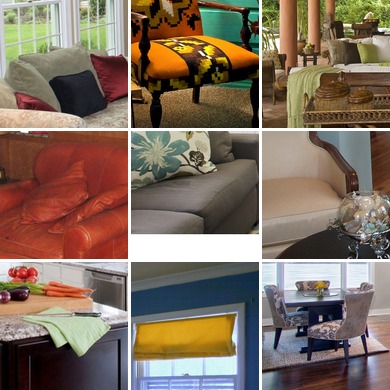} &
      \includegraphics[width=0.105\textwidth]{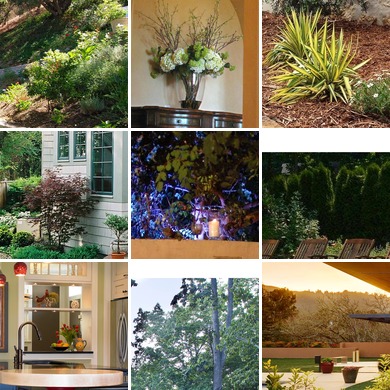} &
      \includegraphics[width=0.105\textwidth]{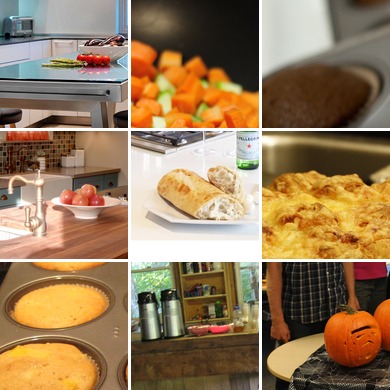} &
      \includegraphics[width=0.105\textwidth]{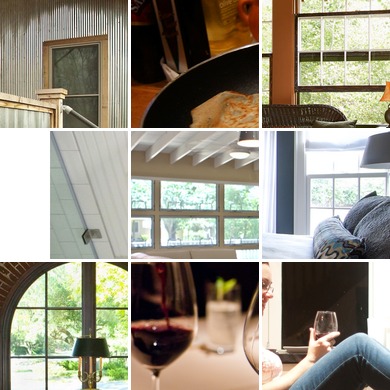} &
      \includegraphics[width=0.105\textwidth]{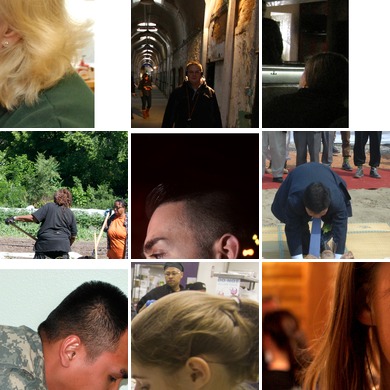}
      \vspace{-4pt}
      \\
      \footnotesize{Brick} &
      \footnotesize{Carpet} &
      \footnotesize{Ceramic} &
      \footnotesize{Fabric} &
      \footnotesize{Foliage} &
      \footnotesize{Food} &
      \footnotesize{Glass} &
      \footnotesize{Hair}
      \\
      \includegraphics[width=0.105\textwidth]{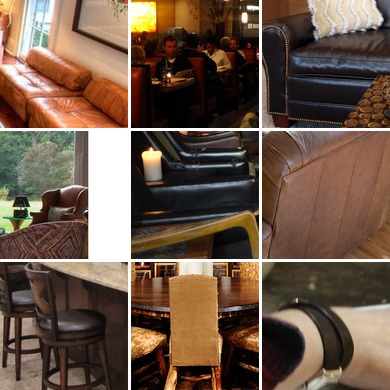} &
      \includegraphics[width=0.105\textwidth]{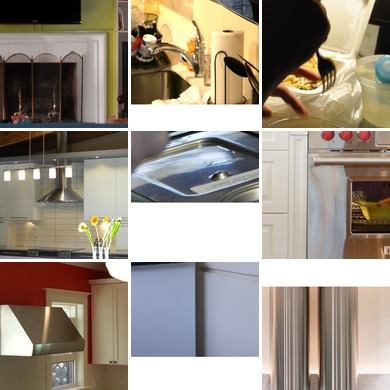} &
      \includegraphics[width=0.105\textwidth]{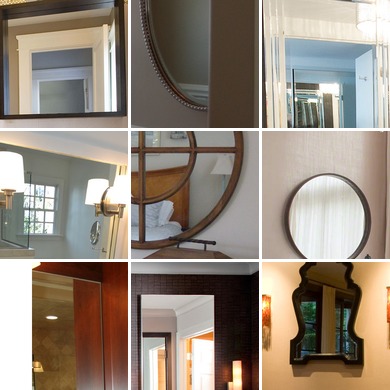} &
      \includegraphics[width=0.105\textwidth]{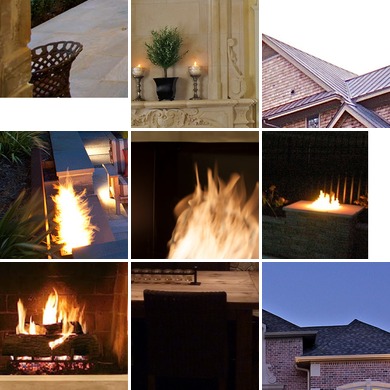} &
      \includegraphics[width=0.105\textwidth]{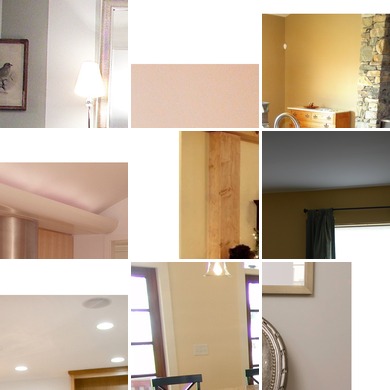} &
      \includegraphics[width=0.105\textwidth]{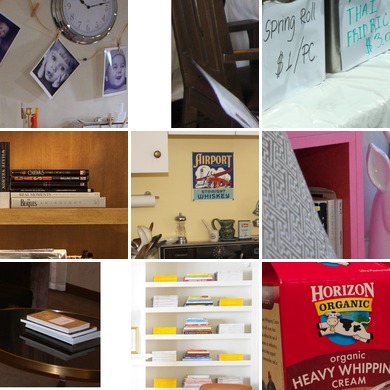} &
      \includegraphics[width=0.105\textwidth]{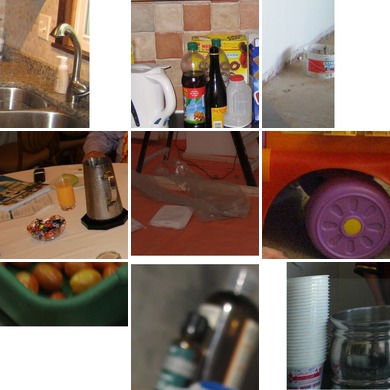} &
      \includegraphics[width=0.105\textwidth]{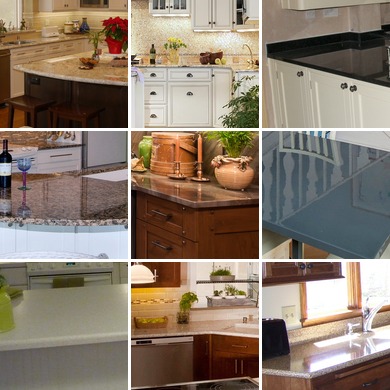}
      \vspace{-4pt}
      \\
      \footnotesize{Leather} &
      \footnotesize{Metal} &
      \footnotesize{Mirror} &
      \footnotesize{Other} &
      \footnotesize{Painted} &
      \footnotesize{Paper} &
      \footnotesize{Plastic} &
      \footnotesize{Pol. stone}
      \vspace{-10pt}
    \end{tabular}
  \end{center}
  \begin{center}
    \begin{tabular}{@{}c@{\hskip 1em}c@{\hskip 1em}c@{\hskip 1em}c@{\hskip 1em}c@{\hskip 1em}c@{\hskip 1em}c@{}}
      \includegraphics[width=0.105\textwidth]{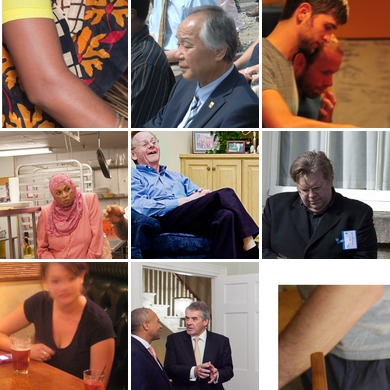} &
      \includegraphics[width=0.105\textwidth]{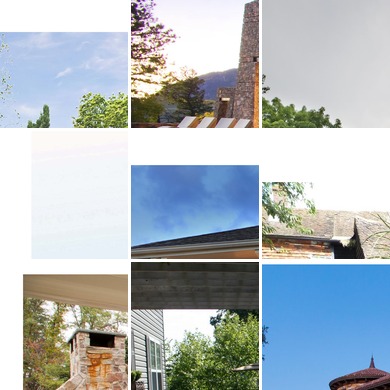} &
      \includegraphics[width=0.105\textwidth]{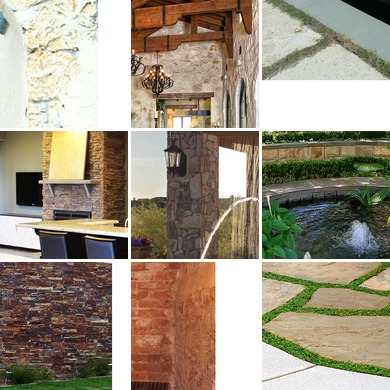} &
      \includegraphics[width=0.105\textwidth]{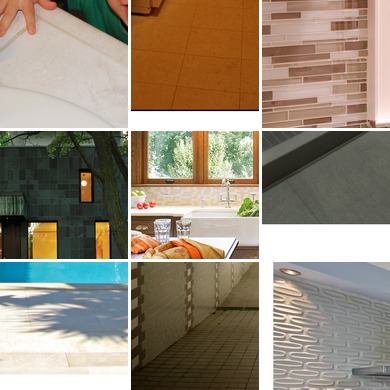} &
      \includegraphics[width=0.105\textwidth]{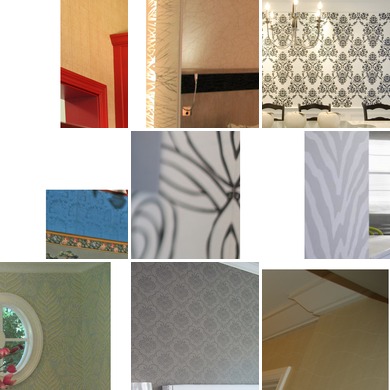} &
      \includegraphics[width=0.105\textwidth]{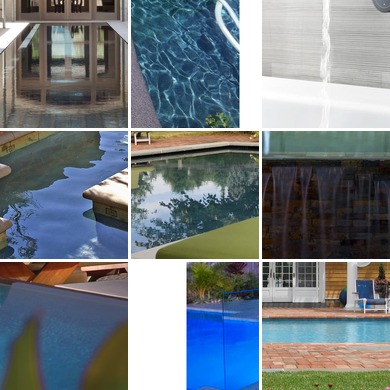} &
      \includegraphics[width=0.105\textwidth]{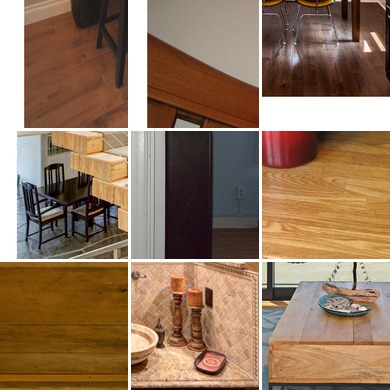}
      \vspace{-4pt}
      \\
      \footnotesize{Skin} &
      \footnotesize{Sky} &
      \footnotesize{Stone} &
      \footnotesize{Tile} &
      \footnotesize{Wallpaper} &
      \footnotesize{Water} &
      \footnotesize{Wood}
    \end{tabular}
  \end{center}
  \vspace{-6pt}
 \caption{%
   Example patches from all 23 categories of the \OStwo (\minc).
   Note that we sample patches so that the
   patch center is the material in question (and not necessarily the
   entire patch).  See Table~\ref{tab:patchcount} for the size of each category.
 }
\label{fig:teaser}
\end{figure*}

A major contribution of our paper is a new, well-sampled material dataset,
called the {\bf \OStwo} (\OStwoshort), with {\bf 3 million} material
samples.  \OStwoshort is more diverse, has more examples in less common
categories, and is much larger than existing datasets.
\OStwoshort draws data from both Flickr images, which include many ``regular''
scenes, as well as Houzz images from professional photographers of staged interiors.
These sources of
images each have different characteristics that together increase the range of
materials that can be recognized.
See Figure~\ref{fig:teaser} for examples of our data.
We make our full dataset available online at \website.

We use this data for material recognition by training different CNN
architectures on this new dataset.  We perform experiments that illustrate the
effect of network architecture, image context, and training data size on
subregions (i.e., patches) of a full scene image.
Further, we build on our patch
classification results and demonstrate simultaneous material recognition and
segmentation of an image by performing dense classification over the image with
a fully connected conditional random field (CRF)
model~\cite{krahenbuhl2013parameter}.  By replacing the fully connected layers
of the CNN with convolutional layers~\cite{overfeat}, the computational burden
is significantly lower than a naive sliding window approach.

In summary, we make two new contributions:
\begin{packed_item}
\item We introduce a new material dataset, \OStwoshort, and 3-stage crowdsourcing pipeline
for efficiently collecting millions of click labels (Section~\ref{sec:clicks}).
\item Our new semantic segmentation method combines a fully-connected CRF with
unary predictions based on CNN learned features (Section~\ref{sec:clickvsseg})
for simultaneous material recognition and segmentation.
\end{packed_item}

\section{Prior Work}

\noindent
{\bf Material Databases.}  Much of the early work on material
recognition focused on classifying specific instances of textures or
material samples. For instance, the CUReT~\cite{dana1999reflectance}
database contains 61 material samples, each captured under 205
different lighting and viewing conditions. This led to research on the
task of instance-level texture or material
classification~\cite{leung2001representing,varma2005statistical}, and
an appreciation of the challenges of building features that
are invariant to pose and illumination. Later, databases with more
diverse examples from each material category began to appear, such as
KTH-TIPS~\cite{hayman2004significance,caputo2005class},
and led explorations of how to
generalize from one example of a material to another---from one sample
of wood to a completely different sample, for instance. Real-world texture
attributes have also recently been explored~\cite{cimpoi2014describing}.

In the domain of categorical material databases, Sharan~\etal released
FMD~\cite{sharan2009material} (described above).  Subsequently, Bell~\etal
released OpenSurfaces~\cite{bell13opensurfaces} which contains over 20,000
real-world scenes labeled with both materials and objects, using a multi-stage
crowdsourcing pipeline.
Because OpenSurfaces images are drawn from consumer photos on Flickr,
material samples have real-world context, in contrast to prior
databases (CUReT, KTH-TIPS, FMD) which feature cropped stand-alone samples.
While OpenSurfaces is a good starting point for a material database, we
substantially expand it with millions of new labels.



\smallskip
\noindent {\bf Material recognition.}
Much prior work on material recognition has focused on the
classification problem (categorizing an image patch into a set of
material categories), often using hand-designed image features. For
FMD, Liu~\etal~\cite{liu2010exploring} introduced reflectance-based edge
features in conjunction with other general image
features. Hu~\etal~\cite{hu2011toward} proposed features based on
variances of oriented gradients. Qi~\etal~\cite{qi2012pairwise}
introduced a pairwise local binary pattern (LBP) feature. Li~\etal~\cite{li2012recognizing} synthesized a dataset based on
KTH-TIPS2 and built a classifier from LBP and dense SIFT.
Timofte~\etal~\cite{timofte2012training} proposed a classification
framework with minimal parameter optimization.  Schwartz and
Nishino~\cite{schwartz2013visual} introduced material traits that
incorporate learned convolutional auto-encoder features. Recently,
Cimpoi~\etal~\cite{cimpoi2014describing} developed a CNN and improved
Fisher vector (IFV) classifier that achieves state-of-the-art results
on FMD and KTH-TIPS2.
Finally, it has been shown that jointly predicting objects and materials can
improve performance~\cite{hu2011toward,zheng2014dense}.

\smallskip
\noindent {\bf Convolutional neural networks.}
While CNNs have been around for a few decades, with early successes
such as LeNet~\cite{lecun89}, they have only recently led to state-of-the-art
results in object classification and detection, leading to enormous
progress.  Driven by the
\new{ILSVRC challenge~\cite{russakovsky2014imagenet},} we have
seen many successful CNN
architectures~\cite{zeilerfergus,overfeat,szegedy2014going,simonyan2014very},
led by the work of Krizhevsky~\etal on their SuperVision \new{(a.k.a. \alexnet)}
network~\cite{krizhevsky2012imagenet}, with more recent architectures
including GoogLeNet~\cite{szegedy2014going}.
In addition to image classification, CNNs are the state-of-the-art for
detection and localization of objects, with recent work including
R-CNNs~\cite{girshick2014rcnn}, Overfeat~\cite{overfeat}, and
VGG~\cite{simonyan2014very}.
Finally, relevant to our goal of per-pixel material segmentation,
Farabet~\etal~\cite{farabet2013learning} use a multi-scale CNN to
predict the class at every pixel in a segmentation.
Oquab~\etal~\cite{oquab2014learning} employ a sliding window approach
to localize patch classification of objects.  We build on this body of
work in deep learning to solve our problem of material recognition and
segmentation.




\section{The \OStwo (\minc)}

We now describe the methodology that went into building our new material
database. Why a new database? We needed a dataset with
the following properties:
\begin{packed_item}
\item{\bf Size}: It should be sufficiently large that learning methods can
  generalize beyond the training set.
\item{\bf Well-sampled}: Rare categories should be represented with a large
  number of examples.
\item{\bf Diversity}: Images should span a wide range of appearances of each
  material in real-world settings.
\item{\bf Number of categories}: It should contain many different
  materials found in the real world.
\end{packed_item}

\subsection{Sources of data} We decided to start with the public, crowdsourced
\OSone dataset~\cite{bell13opensurfaces} as the seed for \minc since
it is drawn from Flickr imagery of everyday, real-world scenes with
reasonable diversity. Furthermore, it has a large number of categories
and the most samples of all prior databases. 

While \OSone data is a good start, it has a few limitations.  Many categories
in \OSone are not well sampled.  While the largest category, {\em wood}, has
nearly 20K samples, smaller categories, such as {\em water}, have only tens of
examples.  This imbalance is due to the way the \OSone dataset was annotated;
workers on Amazon Mechanical Turk (AMT) were free to choose any material
subregion to segment. Workers often gravitated towards certain
common types of materials or salient objects, rather than being
encouraged to label a diverse set of materials. Further, the images come from a
single source (Flickr).

We decided to augment \OSone with substantially more data, especially
for underrepresented material categories, with the initial goal of
gathering at least 10K samples per material category. We decided to
gather this data from another source of imagery, professional photos
on the interior design website Houzz (\url{houzz.com}).  Our
motivation for using this different source of data was that, despite
Houzz photos being more ``staged'' (relative to Flickr photos), they
actually represent a larger variety of materials. For instance,
Houzz photos contain a wide range of types of polished stone.
With these sources of image data, we now describe how we gather material
annotations.



\subsection{Segments, Clicks, and Patches}
\label{sec:clicks}


What specific kinds of material annotations make for a good database?
How should we collect these annotations? The type of annotations to
collect is guided in large part by the tasks we wish to generate
training data for. For some tasks such as scene recognition,
whole-image labels can suffice~\cite{xiao2014sun,zhou2014places}. For
object detection, labeled bounding boxes as in PASCAL are often
used~\cite{everingham10pascal}. For segmentation or scene parsing
tasks, per-pixel segmentations are
required~\cite{russell2008labelme,gould2009decomposing}. Each style of
annotation comes with a cost proportional to its complexity.
For materials, we decided to focus on two problems, guided by prior
work:
\begin{packed_item}
\item {\bf Patch material classification.} Given an image patch, what kind
  of material is it at the center?
\item {\bf Full scene material classification.} Given a full image,
  produce a full per-pixel segmentation and labeling. Also
  known as {\em semantic segmentation} or {\em scene parsing} (but in
  our case, focused on materials). Note that classification can be a
  component of segmentation, e.g., with sliding window approaches.
\end{packed_item}

\smallskip
\noindent{\bf Segments.} \OSone contains material segmentations---carefully
drawn polygons that enclose same-material regions.  To form the basis of MINC,
we selected \OSone segments with high confidence (inter-worker agreement) and
manually curated segments with low confidence, giving a total of 72K shapes.  To
better balance the categories, we manually segmented a few hundred extra samples
for \emph{sky}, \emph{foliage} and \emph{water}.

Since some of the \OSone categories are difficult for humans, we consolidated
these categories.  We found that many AMT workers could not
disambiguate \emph{stone} from \emph{concrete}, \emph{clear plastic} from
\emph{opaque plastic}, and \emph{granite} from \emph{marble}.  Therefore, we
merged these into \emph{stone}, \emph{plastic}, and \emph{polished stone}
respectively.  Without this merging, many ground truth examples in these
categories would be incorrect.  The final list of 23 categories is shown in
Table~\ref{tab:patchcount}.
The
category \emph{other} is different in that it was created by combining various smaller
categories.

\begin{figure}[tb]
\begin{center}
 \includegraphics[width=0.45\textwidth]{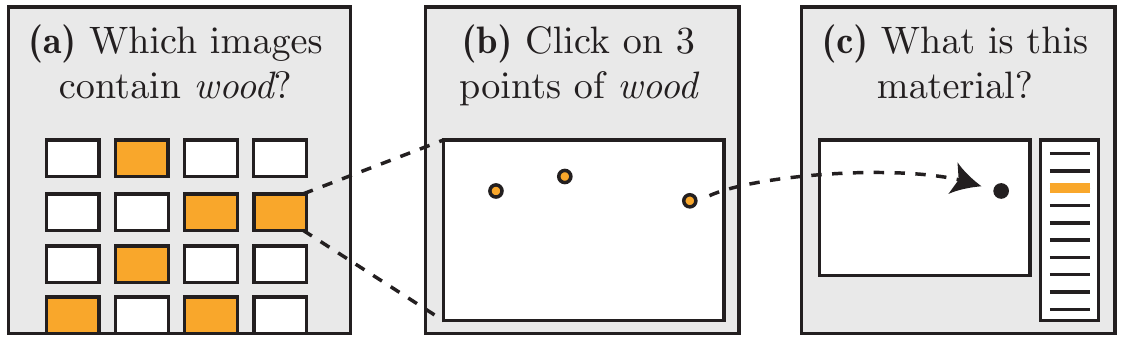}
\end{center}
  \vspace{-8pt}
   \caption{%
     \textbf{AMT pipeline schematic for collecting clicks.}  \textbf{(a)}
     Workers filter by images that contain a certain material, \textbf{(b)}
     workers click on materials, and \textbf{(c)} workers validate click
     locations by re-labeling each point.  Example responses are shown in
     orange.
   }
\label{fig:mturk-overview}
\vspace{-6pt}
\end{figure}

\smallskip
\noindent{\bf Clicks.}
Since we want to expand our dataset to millions of samples, we decided to
augment \OSone segments by collecting {\em clicks}: single points in an image
along with a material label, which are much cheaper and faster to collect.
Figure~\ref{fig:mturk-overview} shows our pipeline for collecting clicks.

Initially, we tried asking workers to click on examples of a given material in a
photo.  However, we found that workers would get frustrated if the material was
absent in too many of the photos.  Thus, we added an initial first
stage where workers
filter out such photos.  To increase the accuracy of our labels, we verify the
click labels by asking different workers to specify the material for each click
without providing them with the label from the previous stage.

To ensure that we obtain high quality annotations and avoid collecting
labels from workers who are not making an effort, we include secret
known answers (sentinels) in the first and third stages, and block
workers with an accuracy below 50\% and 85\% respectively.
We do not use sentinels in the second stage since it would require
per-pixel ground truth labels, and it turned out not to be necessary.  Workers
generally performed all three tasks so we could identify bad workers in the first or
third task.

Material clicks were collected for both OpenSurfaces images and the new Houzz
images.  This allowed us to use labels from OpenSurfaces to generate the
sentinel data; we included 4 sentinels per task.  With this streamlined pipeline
we collected {2,341,473} annotations at an average cost of \$0.00306 per
annotation (stage 1: \$0.02 / 40 images, stage 2: \$0.10 / 50 images, 2, stage
3: \$0.10 / 50 points).

\begin{table}[tb]
  \begin{center}
  \tablesize{%
    \scriptsize{
    \begin{tabular}{|rl|rl|rl|rl|}
      \hline & & & & &\\[-7pt]
      \textbf{Patches} & \textbf{Category} &
      \textbf{Patches} & \textbf{Category} &
      \textbf{Patches} & \textbf{Category} \\
      \hline & & & & &\\[-7pt]
      564,891 & Wood     & 114,085 & Polished stone  & 35,246 & Skin \\
      465,076 & Painted  & 98,891  & Carpet         & 29,616 & Stone \\
      397,982 & Fabric   & 83,644  & Leather        & 28,108 & Ceramic \\
      216,368 & Glass    & 75,084  & Mirror         & 26,103 & Hair \\
      188,491 & Metal    & 64,454  & Brick          & 25,498 & Food \\
      147,346 & Tile     & 55,364  & Water          & 23,779 & Paper \\
      142,150 & Sky      & 39,612  & Other          & 14,954 & Wallpaper \\
      120,957 & Foliage  & 38,975  & Plastic        & & \\
      \hline
    \end{tabular}
    }
  }
  \end{center}
  \vspace{-6pt}
  \caption{\textbf{\minc patch counts by category.} Patches were created from
  both \OSone \segments and our newly collected \emph{clicks}. See
  Section~\ref{sec:clicks} for details.}
  \label{tab:patchcount}
  \vspace{-6pt}
\end{table}

\smallskip
\noindent {\bf Patches.} Labeled segments and clicks form the core of \minc.
For training CNNs and other types of classifiers, it is
useful to have data in the form of fixed-sized patches.  We convert
both forms of data into a unified dataset format: square
image patches.  We use a \emph{patch center} and \emph{patch scale}
(a multiplier of the smaller image dimension) to define the image
subregion that makes a patch. For our patch classification experiments, we use
23.3\% of the smaller image dimension.  Increasing the patch scale provides more
context but reduces the spatial resolution.  Later in Section~\ref{sec:results}
we justify our choice with experiments that vary the patch scale for \alexnet.

We place a patch centered around each click label.
For each \segment, if we were to place a patch
at every interior pixel then we would have a very large and redundant dataset.
Therefore, we Poisson-disk subsample each \segment, separating patch
centers by at least 9.1\% of the smaller image dimension. These \segments
generated {655,201} patches (an average of 9.05 patches per \segment).
In total, we generated {2,996,674} labeled patches from {436,749} images.
%
Patch counts are shown in Table~\ref{tab:patchcount}, and
example patches from various categories are illustrated in Figure~\ref{fig:teaser}.

\begin{figure*}[t]
\begin{center}
   \includegraphics[width=0.99\textwidth]{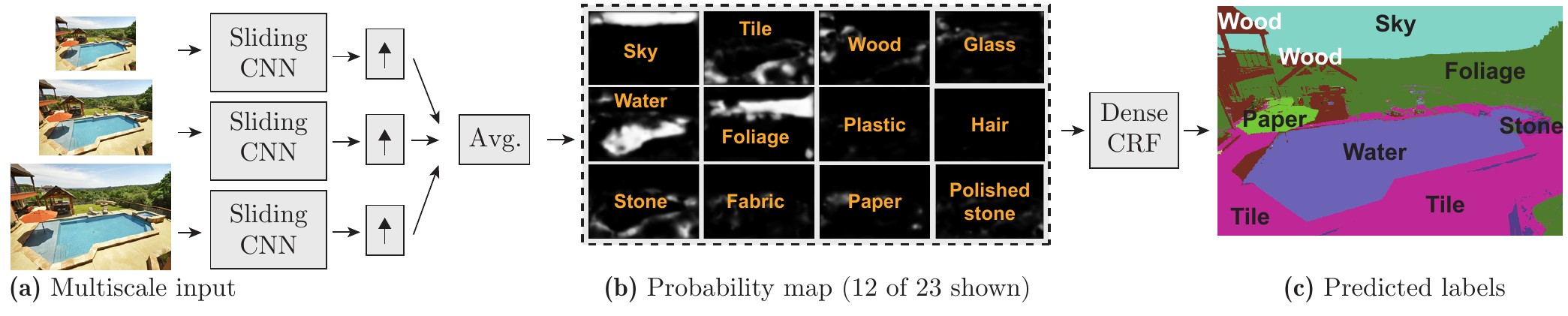}
\end{center}
  \vspace{-10pt}
   \caption{%
     \textbf{Pipeline for full scene material classification.}
     An image \textbf{(a)} is resized to multiple scales $[1/\sqrt{2}, 1,
     \sqrt{2}]$. The same sliding CNN predicts a probability map \textbf{(b)}
     across the image for each scale; the results are upsampled and averaged.
     A fully connected CRF predicts a final label for each pixel \textbf{(c)}.
     This example shows predictions from a single \googlenet converted into a
     sliding CNN (no average pooling).
   }
\label{fig:scene-arch}
\vspace{-10pt}
\end{figure*}

\section{Material recognition in real-world images}

Our goal is to train a system that recognizes the material at every pixel in an
image.
%
We split our training procedure into multiple stages and analyze the performance
of the network at each stage.
First, we train a CNN
that produces a single prediction for a given input patch.
Then, we convert the
CNN into a sliding window and predict materials on a dense
grid across the image.  We do this at multiple scales and average to
obtain a unary term.  Finally, a dense CRF \cite{krahenbuhl2013parameter} combines
the unary term with fully connected pairwise reasoning to
output per-pixel material predictions.  The entire system is depicted in
Figure~\ref{fig:arch-overview}, and described more below.

\subsection{Training procedure}

\minc contains 3 million patches that we split into training, validation and test sets. Randomly splitting would
result in nearly identical patches (e.g., from the same OpenSurfaces
segment) being put in training and test, thus
inflating the test score.  To prevent correlation, we group photos into
clusters of near-duplicates, then assign each cluster to one of
train, validate or test.
We make sure that there are at least 75 \segments of each category in the test set to
ensure there are enough \segments to evaluate segmentation accuracy.
To detect near-duplicates, we compare \alexnet CNN features
computed from each photo (see the supplemental for details).
For exact duplicates, we discard all but one of the copies.

We train all of our CNNs by fine-tuning the network starting from the weights
obtained by training on 1.2 million images from ImageNet (ILSVRC2012).  When
training \alexnet, we use stochastic gradient descent with batchsize 128,
dropout rate 0.5, momentum 0.9, and a base learning rate of $10^{-3}$ that decreases
by a factor of 0.25 every {50,000} iterations.  For \googlenet, we use batchsize
69, dropout 0.4, and learning rate $\alpha_t = 10^{-4} \sqrt{1 - t / 250000}$ for
iteration $t$.

Our training set has a different number of examples per class, so we cycle
through the classes and randomly sample an example from each class.
Failing
to properly balance the examples results in a 5.7\% drop in mean class
accuracy (on the validation set).
Further, since it has been shown to reduce overfitting, we randomly
augment samples by taking crops ($227 \times 227$ out of $256 \times
256$), horizontal mirror flips, spatial scales in the range $[1/\sqrt{2},
\sqrt{2}]$, aspect ratios from 3:4 to 4:3, and amplitude shifts in $[0.95,
1.05]$.
Since we are looking at local regions, we subtract a
per-channel mean (R: 124, G: 117, B: 104) rather than a mean
image~\cite{krizhevsky2012imagenet}.

\subsection{Full scene material classification}
\label{sec:clickvsseg}

Figure~\ref{fig:scene-arch} shows an overview of our method for simultaneously
segmenting and recognizing materials.
Given a CNN that
can classify individual points in the image, we convert it to a sliding window
detector and densely classify a grid across the image.
Specifically, we replace the last fully connected layers with convolutional
layers, so that the network is fully convolutional and can classify images of
any shape.  After conversion, the weights are fixed and not fine-tuned.
With our converted network,
the strides of each layer cause the network to output a prediction every
32 pixels.  We obtain predictions every 16 pixels by shifting the input image by
half-strides (16 pixels).  While this appears to require 4x the computation,
Sermanet \etal~\cite{overfeat} showed that the convolutions can be reused and
only the pool5 through fc8 layers need to be recomputed for the half-stride
shifts.
Adding half-strides resulted in a minor 0.2\% improvement in mean class
accuracy across segments (after applying the dense CRF, described below), and
about the same mean class accuracy at click locations.

The input image is resized so that a patch maps to a 256x256 square. Thus, for a
network trained at patch scale $s$, the resized input has smaller dimension $d = 256/s$. Note that $d$ is inversely proportional to scale, so increased context leads to lower spatial resolution.
We then add padding so that the output probability
map is aligned with the input when upsampled.  We repeat this at 3 different
scales (smaller dimension $d / \sqrt{2}$, $d$, $d \sqrt{2}$), upsample each output
probability map with bilinear interpolation, and average the predictions.  To
make the next step more efficient, we upsample the output to a fixed smaller
dimension of $550$.

We then use the dense CRF of~\krahenbuhl
\etal~\cite{krahenbuhl2013parameter} to predict a label at every
pixel,
using the following energy:
%
\vspace{-2pt}
\begin{align}
    E(x \given \mathbf{I}) &=
      \sum_i \psi_i(x_i) +
      \sum_{i < j} \psi_{ij}(x_i, x_j)
  \label{eq:dense-crf}
  \\
  \psi_{i}(x_i) &= -\log p_i(x_i) \\
  \psi_{ij}(x_i, x_j) &= w_p \ \delta(x_i \ne x_j) \ k(\mathbf{f}_i - \mathbf{f}_j)
  \label{eq:dense-crf-pairwise}
\end{align}
where $\psi_i$ is the unary energy (negative log of the aggregated softmax
probabilities) and $\psi_{ij}$ is the pairwise term that connects every pair of
pixels in the image.  We use a single pairwise term with a Potts label compatibility term
$\delta$ weighted by $w_p$ and unit Gaussian kernel $k$.
For the features $\mathbf{f}_i$, we convert the RGB image to
$L^\text{*}a^\text{*}b^\text{*}$ and use color $(I^L_i, I^a_i, I^b_i)$ and
position $(p^x,p^y)$ as pairwise features for each pixel:
  $\mathbf{f}_i =
  \left[
    {p^x_i \over \theta_p \, d},
    {p^y_i \over \theta_p \, d},
    {I^L_i \over \theta_L},
    {I^a_i \over \theta_{ab}},
    {I^b_i \over \theta_{ab}}
  \right]$,
where $d$ is the smaller image dimension.
Figure~\ref{fig:scene-arch} shows an example unary term $p_i$ and the resulting
segmentation $x$.



\begin{figure}[t]
\begin{center}
   \includegraphics[height=0.43\linewidth]{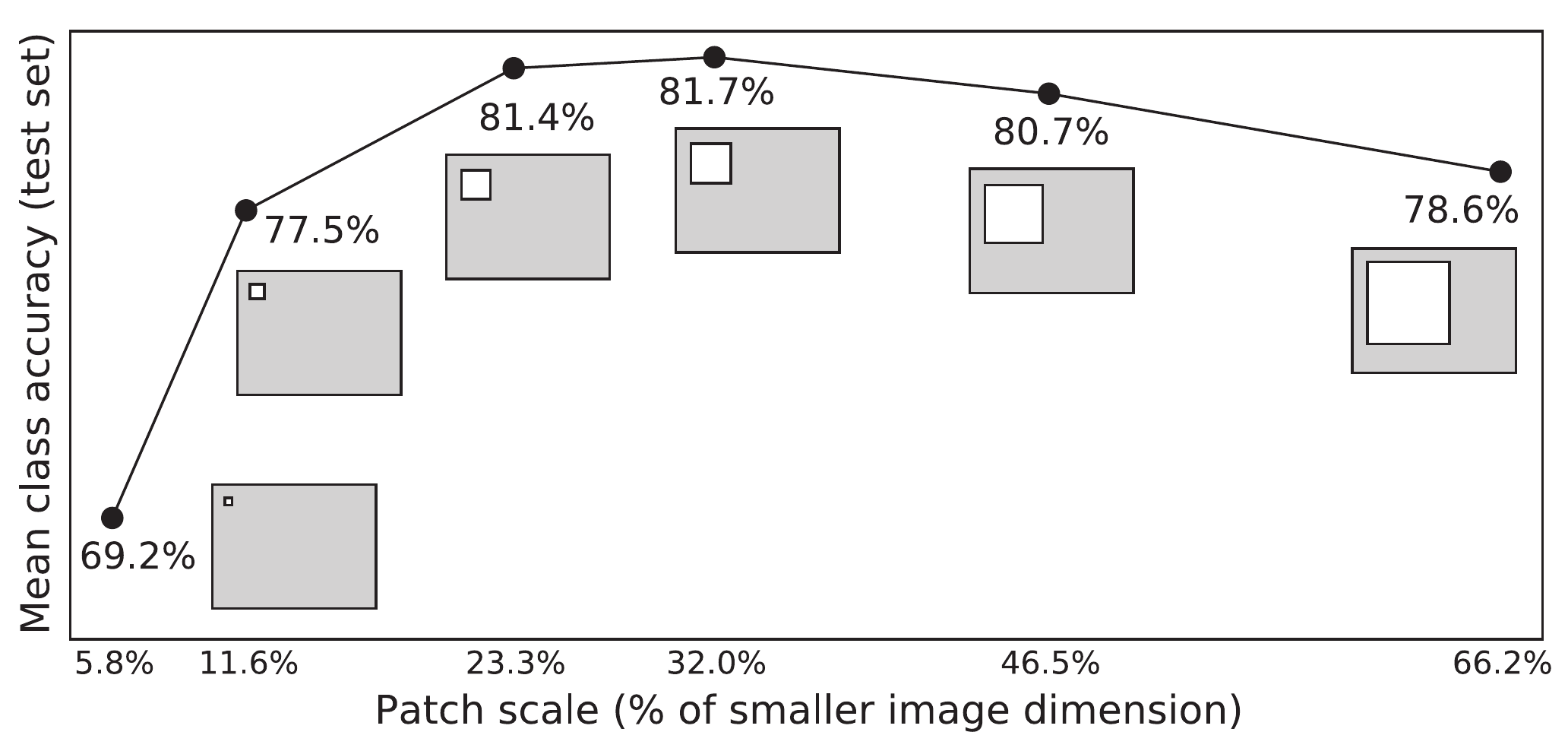}
\end{center}
\vspace{-10pt}
\caption{\textbf{Varying patch scale}.  We train/test patches of different
scales (the patch locations do not vary).  The optimum is a trade-off between
context and spatial resolution. CNN: \alexnet.}
\label{fig:patchsize}
\vspace{-4pt}
\end{figure}

\section{Experiments and Results}
\label{sec:results}

\subsection{Patch material classification}
\label{sec:results-patch}

\begin{table}
\begin{center}
\tablesize{%
  \begin{tabular}{|l|l|l|}
  \hline
  \textbf{Architecture} & \textbf{Validation} & \textbf{Test} \\
  \hline
  \alexnet~\cite{krizhevsky2012imagenet}    & 82.2\% & 81.4\% \\
  \googlenet~\cite{szegedy2014going}  & \textbf{85.9\%} & \textbf{85.2\%} \\
  \vgg~\cite{simonyan2014very}        & 85.6\% & 84.8\% \\
  \hline
  \end{tabular}
}
\end{center}
\vspace{-4pt}
\caption{
  \textbf{Patch material classification results.}
  Mean class accuracy for different CNNs trained on MINC.
  See Section~\ref{sec:results-patch}.}
\label{tab:netarch}
\vspace{-6pt}
\end{table}

\begin{table}[tb]
  \begin{center}
  \tablesize{%
    \scriptsize{
    \begin{tabular}{|r@{\hskip 0.55em}l|r@{\hskip 0.55em}l|r@{\hskip 0.55em}l|r@{\hskip 0.55em}l|r@{\hskip 0.55em}l|}
      \hline & & & & & & &\\[-7pt]
      Sky     & 97.3\%  & Food        & 90.4\% & Wallpaper  & 83.4\% & Glass   & 78.5\% \\
      Hair    & 95.3\%  & Leather     & 88.2\% & Tile       & 82.7\% & Fabric  & 77.8\% \\
      Foliage & 95.1\%  & Other       & 87.9\% & Ceramic    & 82.7\% & Metal   & 77.7\% \\
      Skin    & 93.9\%  & Pol. stone  & 85.8\% & Stone      & 82.7\% & Mirror  & 72.0\% \\
      Water   & 93.6\%  & Brick       & 85.1\% & Paper      & 81.8\% & Plastic & 70.9\% \\
      Carpet  & 91.6\%  & Painted     & 84.2\% & Wood       & 81.3\% &         & \\
      \hline
    \end{tabular}
    }
  }
  \end{center}
  \vspace{-6pt}
  \caption{\textbf{Patch test accuracy by category.} CNN: \googlenet.
  See the supplemental material for a full confusion matrix.}
  \label{tab:patchcategory}
  \vspace{-6pt}
\end{table}

In this section, we evaluate the effect of many different design
decisions for training methods for material classification and
segmentation, including various CNN architectures, patch sizes, and
amounts of data.

\begin{figure}[t]
\begin{center}
  \begin{tabular}{@{}c@{\hskip 0.3em}c@{\hskip 0.3em}c@{\hskip 0.3em}c@{\hskip 0.3em}c@{}}
    \rotatebox{90}{\quad\ \tablesize{Correct}} &
    \includegraphics[width=0.11\textwidth]{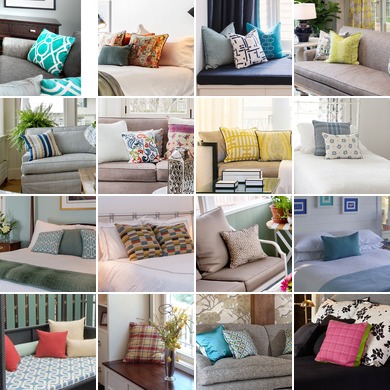}
    &
    \includegraphics[width=0.11\textwidth]{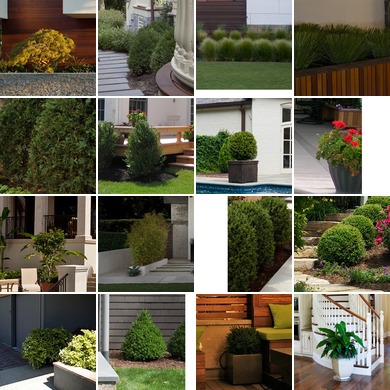}
    &
    \includegraphics[width=0.11\textwidth]{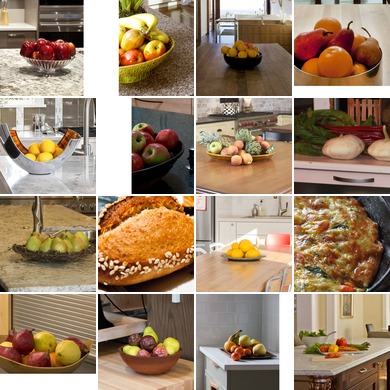}
    &
    \includegraphics[width=0.11\textwidth]{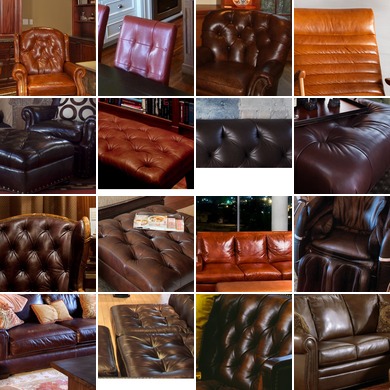}
    \vspace{-2pt}
    \\
    &
    \tablesize{Fabric $(99\%)$} &
    \tablesize{Foliage $(99\%)$} &
    \tablesize{Food $(99\%)$} &
    \tablesize{Leather $(99\%)$}
    \vspace{2pt}
    \\
    \rotatebox{90}{\quad\ \tablesize{Correct}} &
    \includegraphics[width=0.11\textwidth]{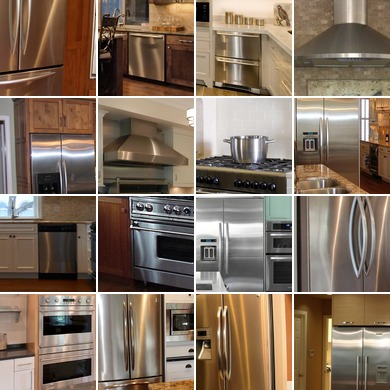}
    &
    \includegraphics[width=0.11\textwidth]{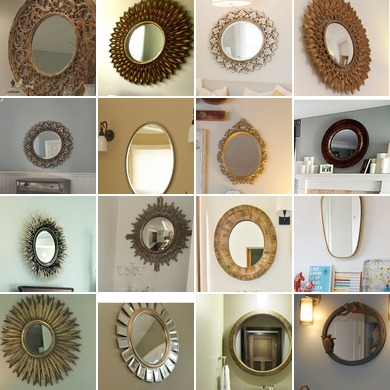}
    &
    \includegraphics[width=0.11\textwidth]{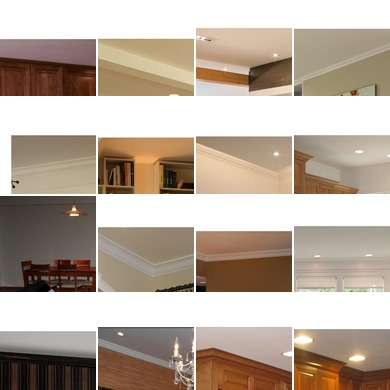}
    &
    \includegraphics[width=0.11\textwidth]{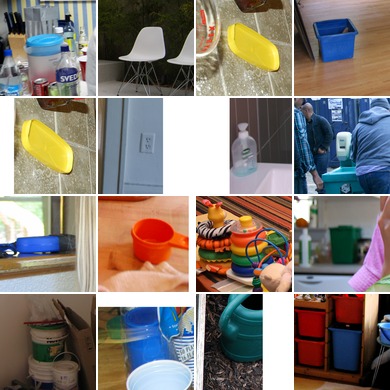}
    \vspace{-2pt}
    \\
    &
    \tablesize{Metal $(99\%)$} &
    \tablesize{Mirror $(99\%)$} &
    \tablesize{Painted $(99\%)$} &
    \tablesize{Plastic $(99\%)$}
    \vspace{1pt}
    \\ \hline \\ [-2ex]
    \rotatebox{90}{\quad\ \tablesize{Incorrect}} &
    \includegraphics[width=0.11\textwidth]{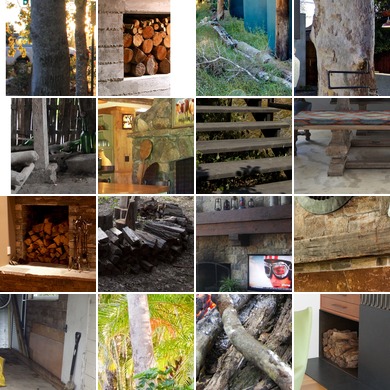}
    &
    \includegraphics[width=0.11\textwidth]{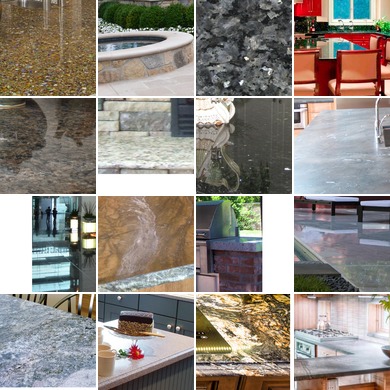}
    &
    \includegraphics[width=0.11\textwidth]{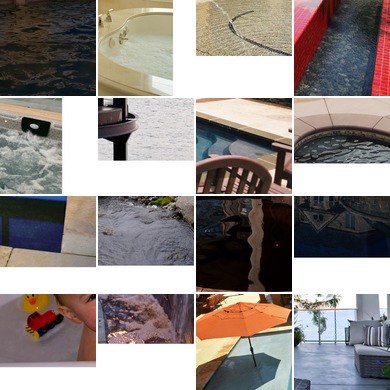}
    &
    \includegraphics[width=0.11\textwidth]{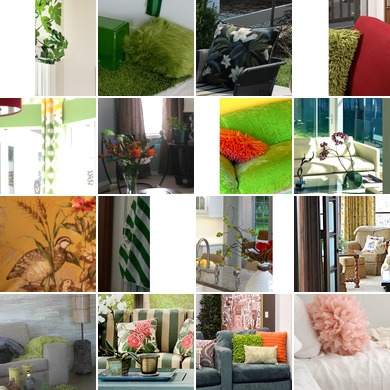}
    \vspace{-2pt}
    \\
    &
    \tablesize{\textbf{T}: Wood} &
    \tablesize{\textbf{T}: Polished stone} &
    \tablesize{\textbf{T}: Water} &
    \tablesize{\textbf{T}: Fabric } \\
    &
    \tablesize{\textbf{P}: Stone $(90\%)$} &
    \tablesize{\textbf{P}: Water $(35\%)$} &
    \tablesize{\textbf{P}: Carpet $(27\%)$} &
    \tablesize{\textbf{P}: Foliage $(67\%)$}
  \end{tabular}
\end{center}
\vspace{-6pt}
\caption{%
  \textbf{High confidence predictions.}
  Top two rows: correct predictions.  Bottom row: incorrect predictions
  (\textbf{T}: true, \textbf{P}: predicted).  Percentages indicate confidence
  (the predictions shown are at least this confident). CNN: \googlenet.
}
\label{fig:patchvisualcm}
\vspace{-6pt}
\end{figure}

\smallskip
\noindent{\bf CNN Architectures.}
Our ultimate goal is full material segmentation, but we are also
interested in exploring which CNN architectures give the best results for classifying single
patches.  Among the networks and parameter variations we tried we found
the best performing networks were \alexnet~\cite{krizhevsky2012imagenet},
\vgg~\cite{simonyan2014very} and \googlenet~\cite{szegedy2014going}.  \alexnet
and \googlenet are re-implementations by BVLC~\cite{caffe}, and \vgg is configuration D (a 16
layer network) of~\cite{simonyan2014very}.
All models were obtained from the Caffe Model Zoo~\cite{caffe}.
Our experiments use \alexnet for evaluating material classification design
decisions and combinations of \alexnet and \googlenet for evaluating material
segmentation.  Tables~\ref{tab:netarch} and
\ref{tab:patchcategory} summarize patch material classification results on our dataset. Figure~\ref{fig:patchvisualcm} shows correct and incorrect predictions made with high confidence.

\begin{figure}[t]
\begin{center}
   \includegraphics[width=0.45\textwidth]{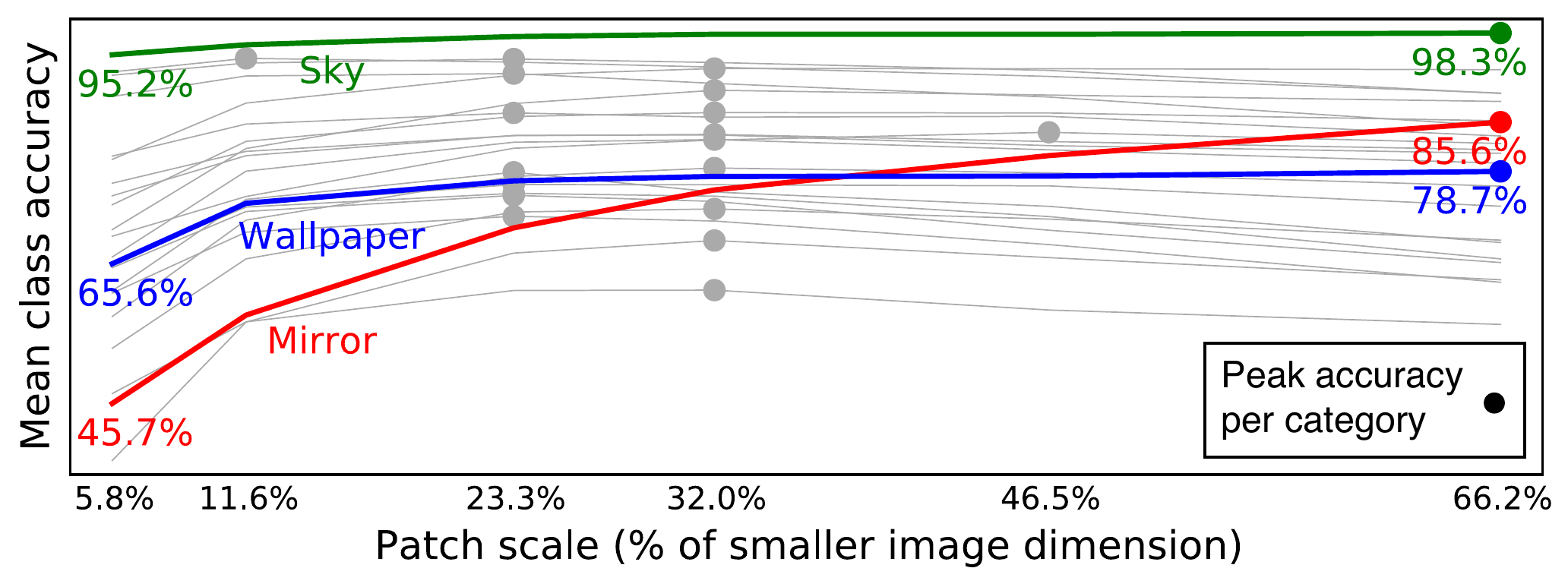}
\end{center}
\vspace{-10pt}
  \caption{\textbf{Accuracy vs patch scale by category}. Dots: peak accuracy for each category; colored lines:
  \emph{sky}, \emph{wallpaper}, \emph{mirror}; gray lines: other categories.  CNN: \alexnet. While
  most materials are optimally recognized at 23.3\% or 32\% patch scale,
  recognition of \emph{sky}, \emph{wallpaper} and \emph{mirror} improve with increasing context.}
\label{fig:categoryaccuracy}
\vspace{-6pt}
\end{figure}

\smallskip
\noindent{\bf Input patch scale.}
To classify a point
in an image we must decide how much context to include around
it. The context, expressed as a fraction of image size, is the patch scale.  A priori, it is not clear which scale is best since small patches have
better spatial resolution, but large patches have more contextual
information. Holding patch centers fixed we varied scale and
evaluated classification accuracy with \alexnet.
Results and a visualization of patch scales are shown in
Figure~\ref{fig:patchsize}. Scale 32\% performs the best. Individual categories had peaks at middle scale with some
exceptions; we find that \emph{mirror}, \emph{wallpaper} and \emph{sky} improve with increasing
context (Figure~\ref{fig:categoryaccuracy}).
We used 23.3\% (which has nearly the same accuracy but higher spatial resolution) for our experiments.


\begin{figure}[t]
\begin{center}
   \includegraphics[height=0.43\linewidth]{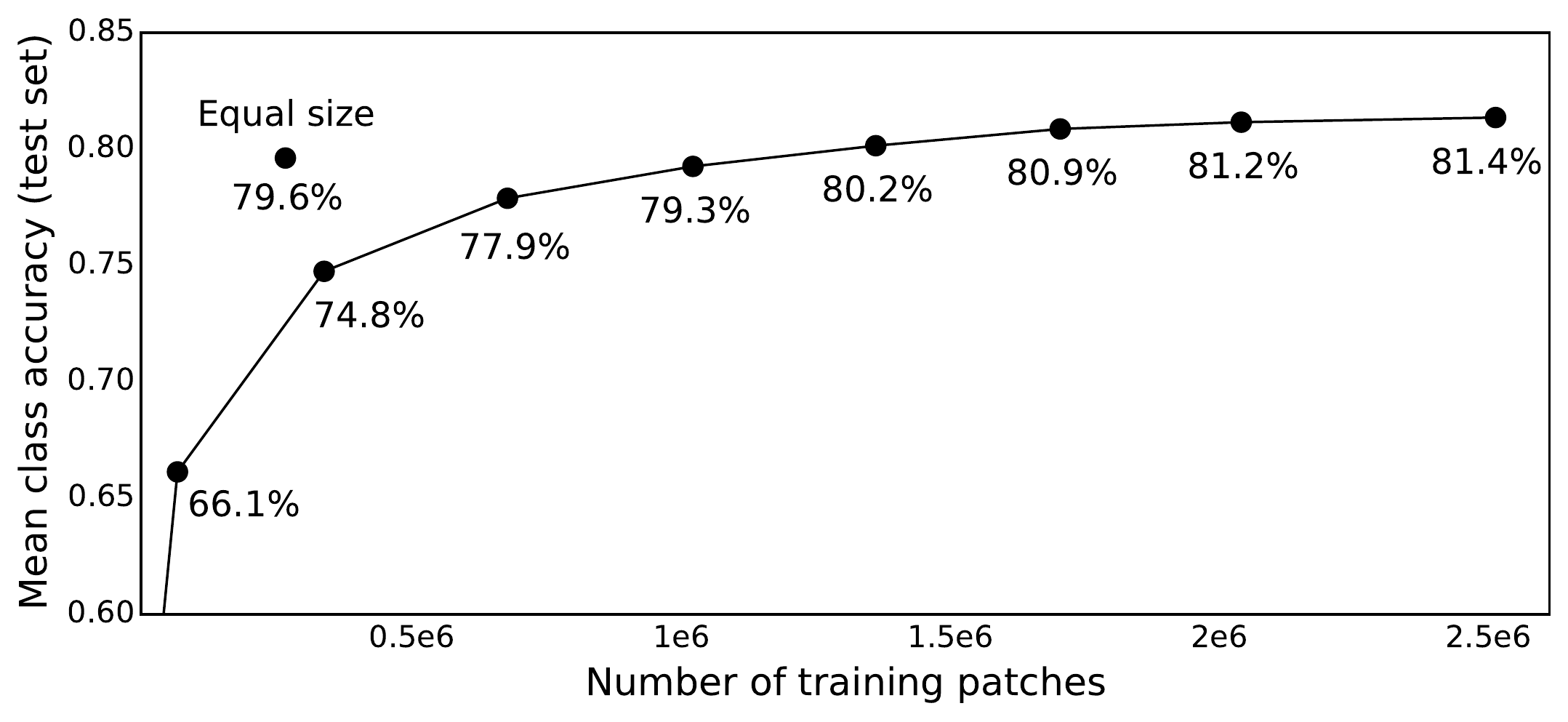}
\end{center}
\vspace{-10pt}
\caption{\textbf{Varying database size.}
    Patch accuracy when trained on random subsets of \minc.
    \emph{Equal size} is using equal samples per category (size
   determined by smallest category).  CNN: \alexnet.
 }
\label{fig:scatter}
\vspace{-4pt}
\end{figure}

\begin{figure*}[t]
\begin{center}
  \begin{minipage}{0.8\textwidth}
    \begin{tabular}{@{}c@{\hskip 0.3em}c@{\hskip 0.3em}c@{\hskip 0.3em}c@{\hskip 0.3em}c@{\hskip 0.3em}c@{}}
     \includegraphics[height=0.1595\textwidth]{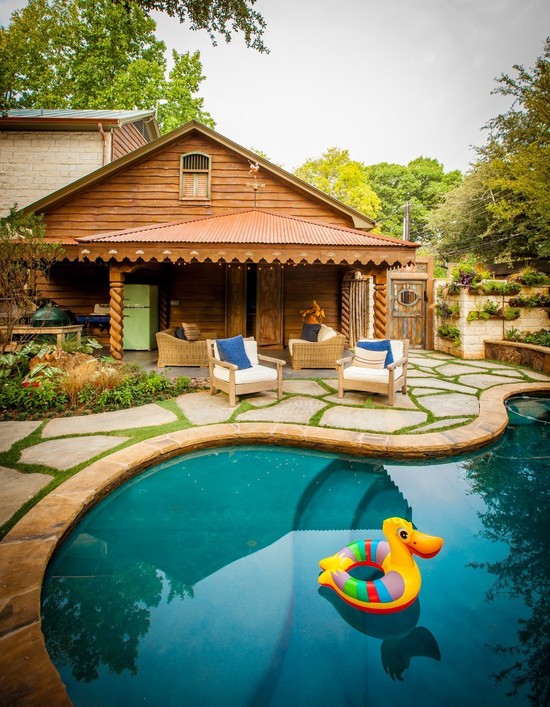} &
     \includegraphics[height=0.1595\textwidth]{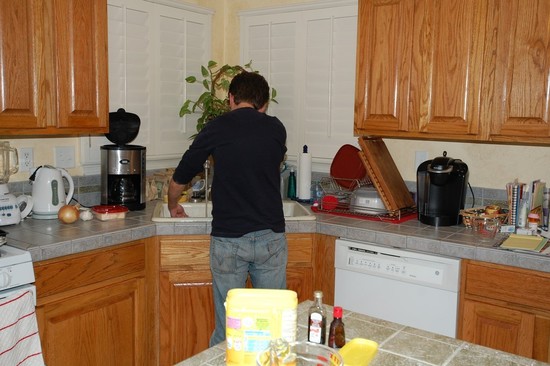} &
     \includegraphics[height=0.1595\textwidth]{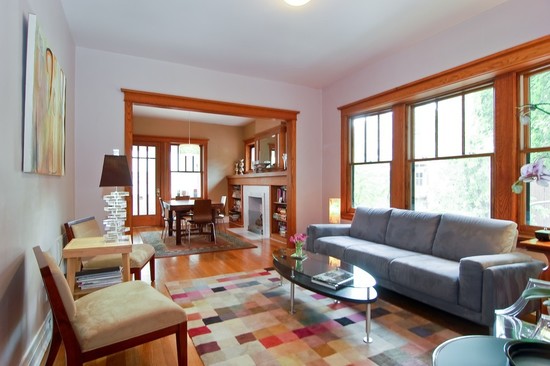} &
     \includegraphics[height=0.1595\textwidth]{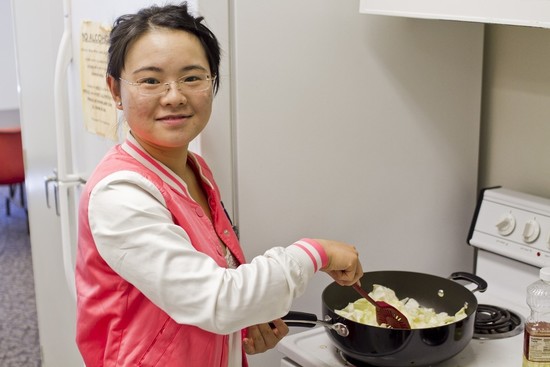} &
     \includegraphics[height=0.1595\textwidth]{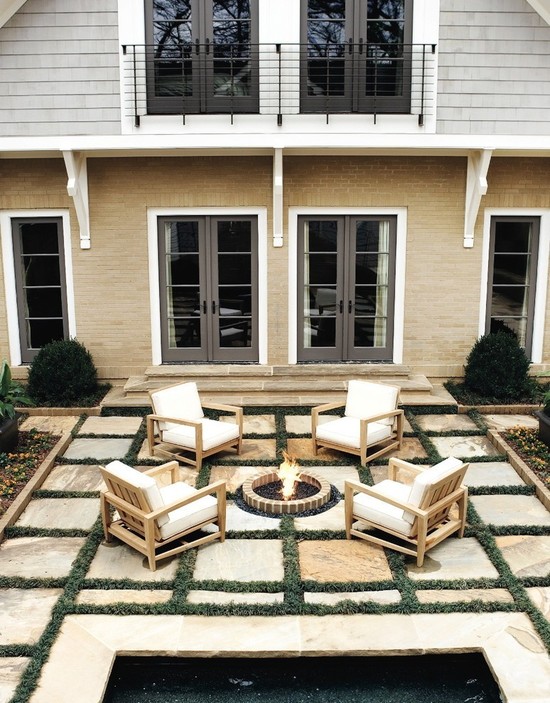}
     \\
     \includegraphics[height=0.1595\textwidth]{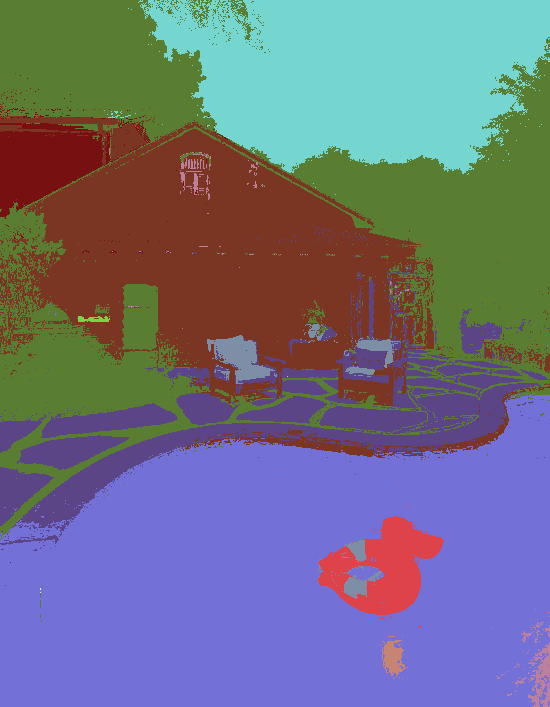} &
     \includegraphics[height=0.1595\textwidth]{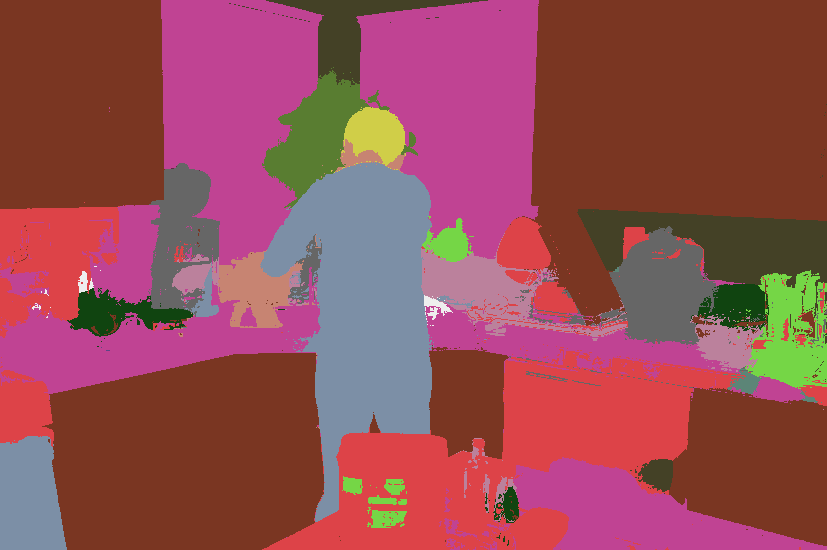} &
     \includegraphics[height=0.1595\textwidth]{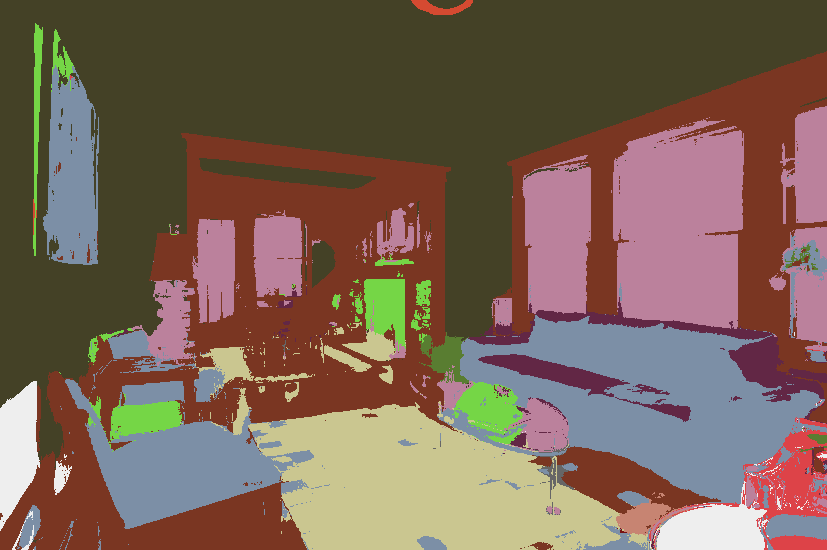} &
     \includegraphics[height=0.1595\textwidth]{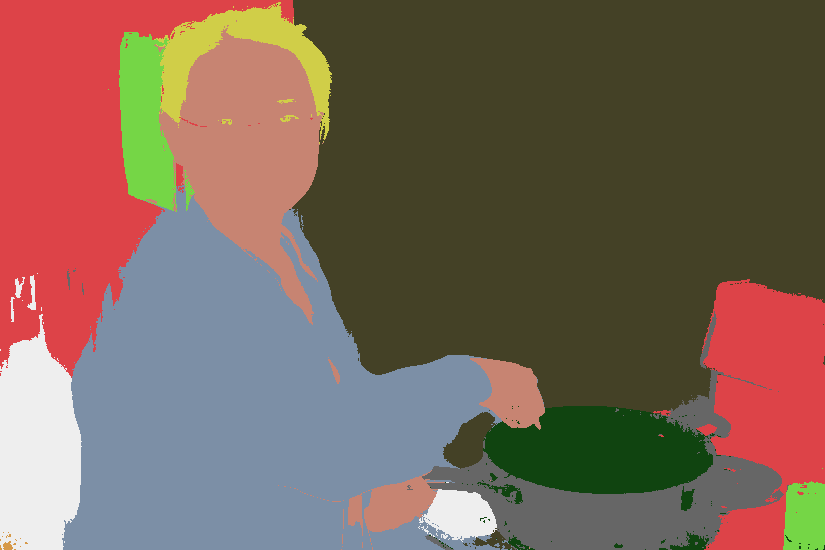} &
     \includegraphics[height=0.1595\textwidth]{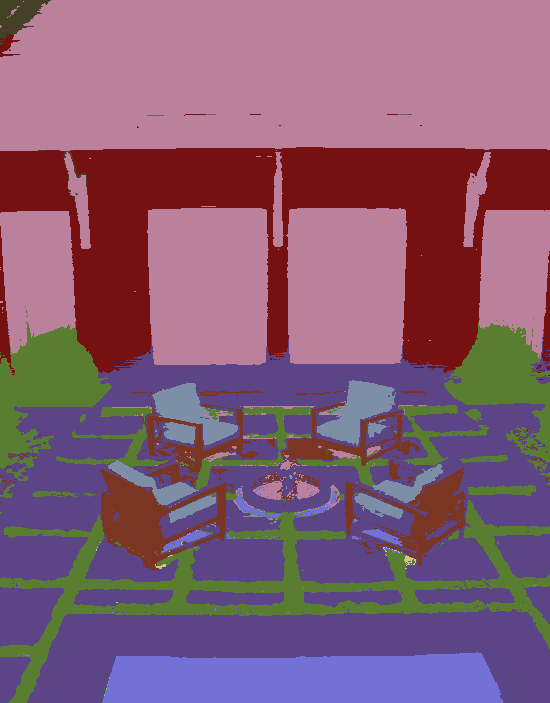}
    \end{tabular}
  \end{minipage}
  \begin{minipage}{0.19\textwidth}
     \includegraphics[width=1.05\textwidth]{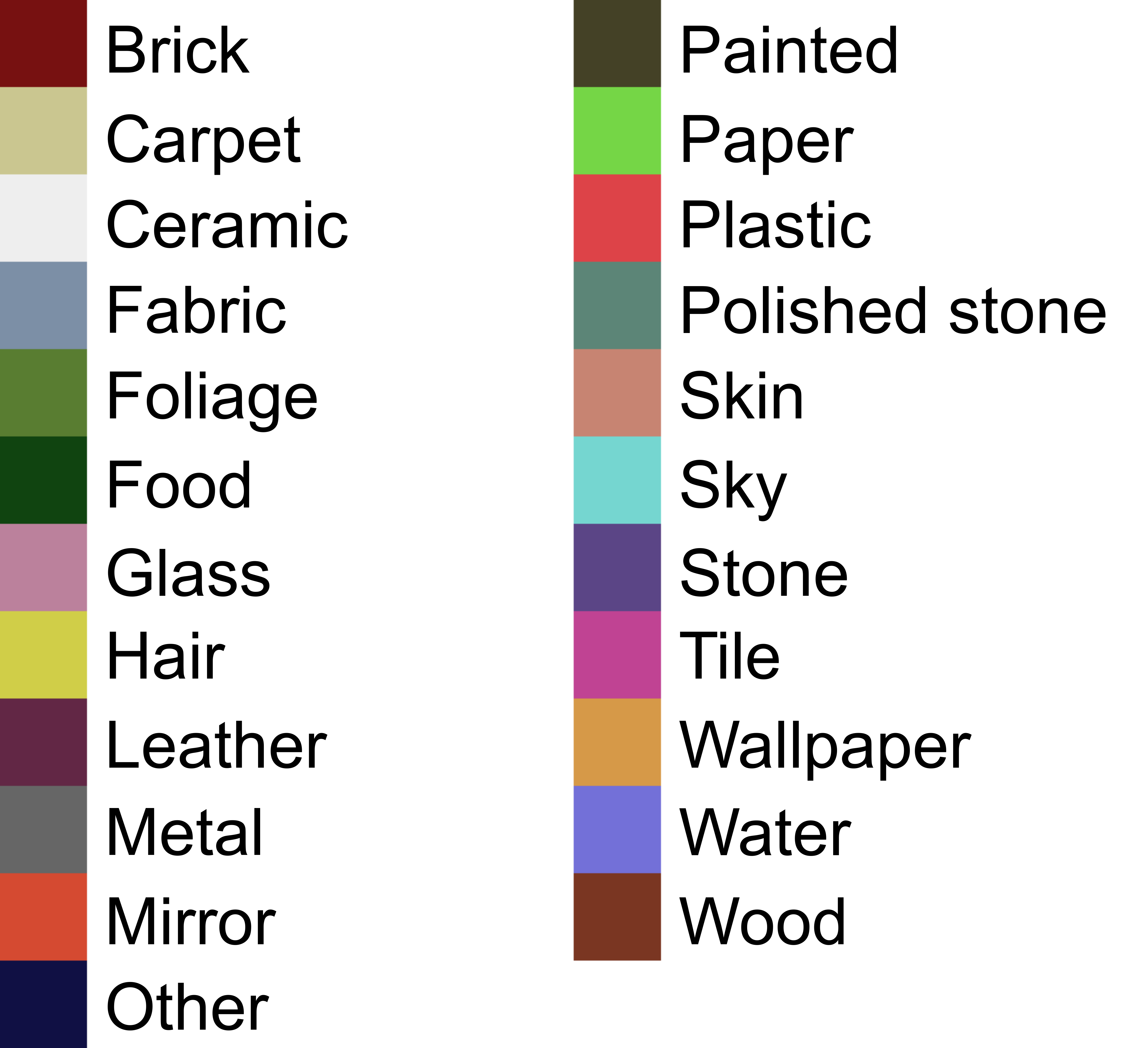}
  \end{minipage}
\end{center}
  \vspace{-6pt}
  \caption{%
    \textbf{Full-scene material classification examples:} high-accuracy test set
    predictions by our method.  CNN: \googlenet (with the average pooling layer
    removed).
    Right: legend for material colors.
    See Table~\ref{tab:sceneaccuracy} for quantitative evaluation.
  }
  \label{fig:sceneresults}
  \vspace{-10pt}
\end{figure*}

\smallskip
\noindent{\bf Dataset size.}
\label{sec:datasetsize} To measure the
effect of size on patch classification accuracy we trained \alexnet with
patches from randomly sampled subsets of all 369,104
training images and tested
on our full test set (Figure~\ref{fig:scatter}). As expected, using more data improved performance. In addition,
we still have not saturated performance with 2.5 million training patches;
even higher accuracies may be possible with more training data (though
with diminishing returns).



\smallskip
\noindent{\bf Dataset balance.}
Although we've shown that more data is better we also find that a balanced
dataset is more effective. We trained \alexnet with all patches of our smallest
category (\emph{wallpaper}) and randomly sampled the larger categories (the
largest, \emph{wood}, being 40x larger) to be equal size.  We then measured mean
class accuracy on the same full test set. As shown in Figure~\ref{fig:scatter},
``Equal size'' is more accurate than a dataset of the same size and just 1.7\%
lower than the full training set (which is 9x larger).
This result further demonstrates the value of building up datasets in a
balanced manner, focusing on expanding the smallest, least common categories.

\begin{figure}[t]
\begin{center}
  \begin{tabular}{@{}c@{\hskip 0.3em}c@{\hskip 0.3em}c@{}}
   \includegraphics[width=0.32\linewidth]{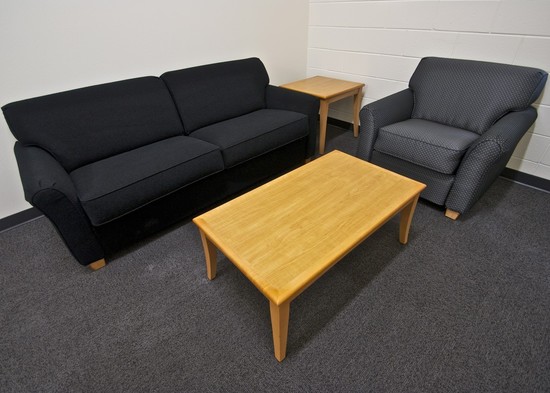} &
   \includegraphics[width=0.32\linewidth]{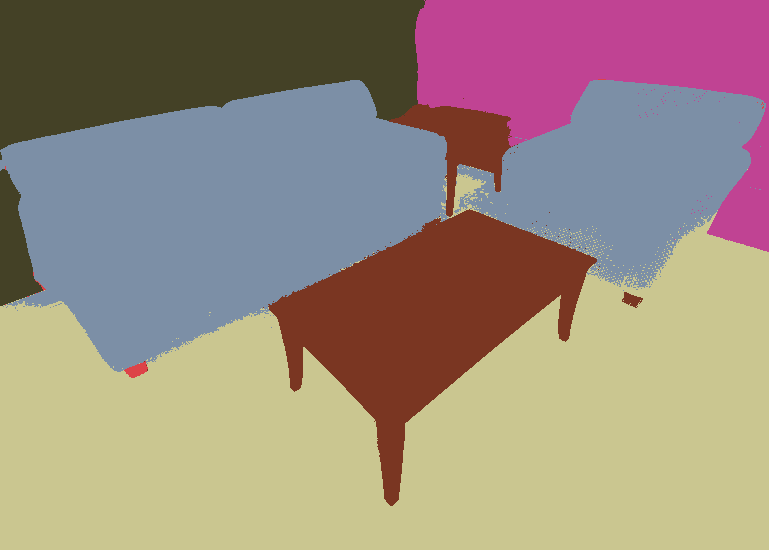} &
   \includegraphics[width=0.32\linewidth]{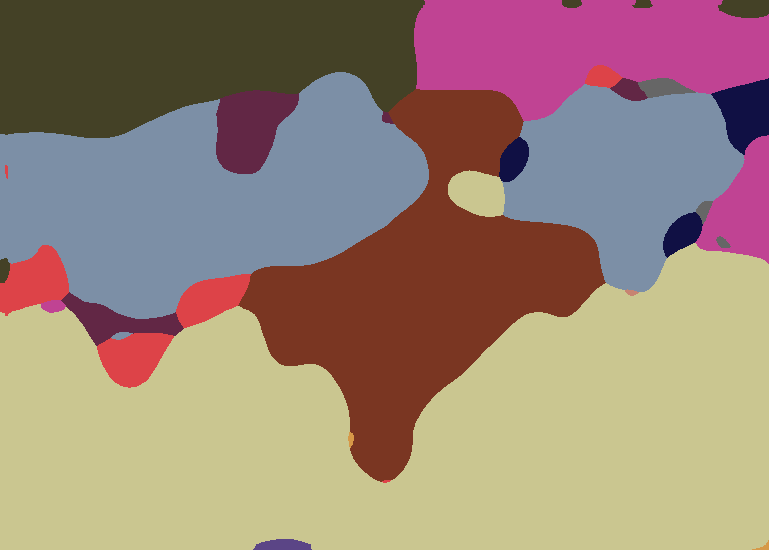}
   \\
   \includegraphics[width=0.32\linewidth]{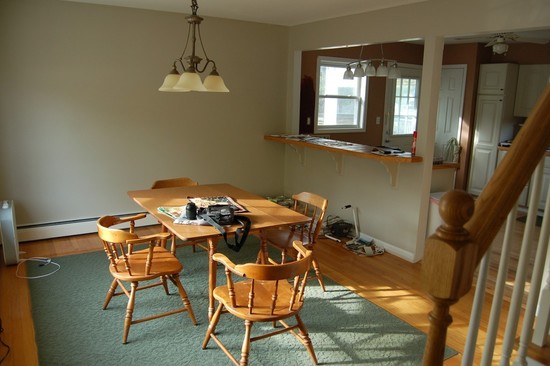} &
   \includegraphics[width=0.32\linewidth]{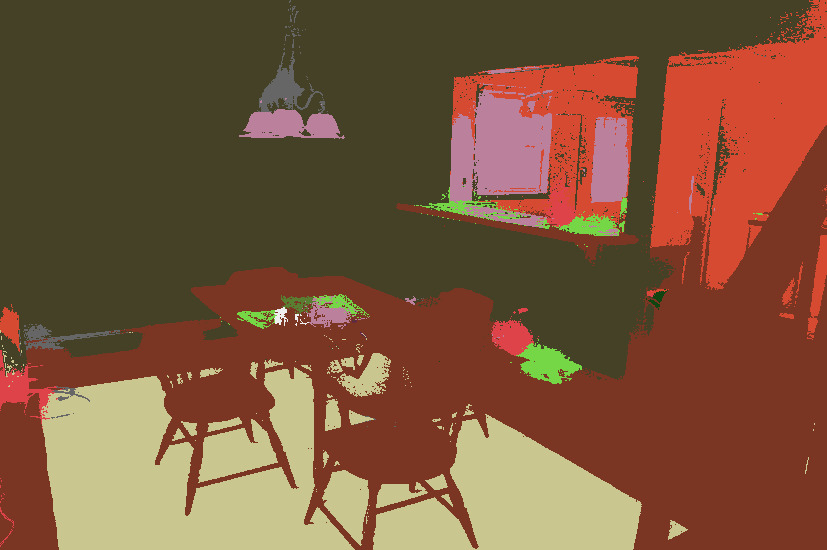} &
   \includegraphics[width=0.32\linewidth]{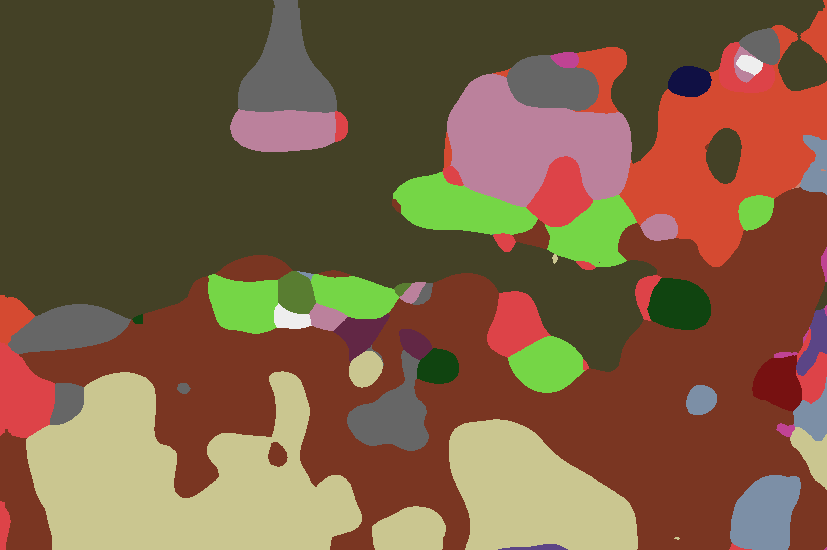}
   \vspace{-2pt}
   \\
   \footnotesize{Input image} &
   \footnotesize{\textbf{(a)} Labels from CRF} &
   \footnotesize{\textbf{(b)} Labels from CRF}
   \vspace{-2pt} \\
   \footnotesize{(test set)} &
   \footnotesize{trained on segments} &
   \footnotesize{trained on clicks}
 \end{tabular}
\end{center}
  \vspace{-6pt}
   \caption{%
     \textbf{Optimizing for click accuracy leads to sloppy boundaries.}
     In \textbf{(a)}, we optimize for mean class accuracy across segments,
     resulting in high quality boundaries.
     In \textbf{(b)}, we optimize for mean class accuracy at click locations.
     Since the clicks are not necessarily close to object boundaries, there is
     no penalty for sloppy boundaries.
     CNN: \googlenet (without average pooling).
   }
  \label{fig:withoutcrf}
  \vspace{-6pt}
\end{figure}

\subsection{Full scene material segmentation}\label{text:fullscene}

The full test set for our patch dataset contains {41,801} photos, but most of
them contain only a few labels.  Since we want to evaluate the per-pixel
classification performance, we select a subset of {5,000} test photos such that
each photo contains a large number of segments and clicks, and small categories
are well sampled.  We greedily solve for the best such set of photos.  We
similarly select {2,500} of {25,844} validation photos.  Our splits for all
experiments are included online with the dataset.
To train the CRF for our model, we try various parameter settings ($\theta_p$,
$\theta_{ab}$, $\theta_L$, $w_p$) and select the model that performs best on the
validation set.  In total, we evaluate 1799 combinations of CNNs and CRF
parameters.  See the supplemental material for a detailed breakdown.

\begin{table}
\begin{center}
\tablesize{%
  \begin{tabular}{|l@{\hskip 1em}l|c@{\hskip 0.5em}c||c@{\hskip 1em}c|}
  \hline
  \multicolumn{2}{|c|}{%
    \multirow{2}{*}{Architecture}
  } &
  \multicolumn{2}{c||}{\textbf{(a)} Segments only} &
  \multicolumn{2}{c|}{\textbf{(b)} Clicks only}
  \\
  \cline{3-6}
   & & Class & Total & Class & Total \\
  \hline
  \alexnet & Scale: 11.6\% & 64.3\% & 72.6\%  & 79.9\% & 77.2\%  \\ 
  \alexnet & Scale: 23.3\% & 69.6\% & 76.6\%  & \textbf{83.3\%} & \textbf{81.1\%}  \\ 
  \alexnet & Scale: 32.0\% & \textbf{70.1\%} & \textbf{77.1\%}  & 83.2\% & 80.7\%  \\ 
  \alexnet & Scale: 46.5\% & 69.6\% & 75.4\%  & 80.8\% & 77.7\%  \\ 
  \alexnet & Scale: 66.2\% & 67.7\% & 72.0\%  & 77.2\% & 72.6\%  \\ 
  \hline
  \googlenet & 7x7 avg. pool & 64.4\% & 71.6\%  & 63.6\% & 63.4\%  \\ 
  \googlenet & 5x5 avg. pool & 67.6\% & 74.6\%  & 70.9\% & 69.8\%  \\ 
  \googlenet & 3x3 avg. pool & \textbf{70.4\%} & 77.7\%  & 76.1\% & 74.7\%  \\ 
  \googlenet & No avg. pool & \textbf{70.4\%} & \textbf{78.8\%}  & \textbf{79.1\%} & \textbf{77.4\%}  \\ 
  \hline
  Ensemble & 2 CNNs &  \textbf{73.1\%} & \textbf{79.8\%}  & 84.5\% & 83.1\%  \\
  Ensemble & 3 CNNs &  \textbf{73.1\%} & 79.3\%  & \textbf{85.9\%} & \textbf{83.5\%}  \\
  Ensemble & 4 CNNs &  72.1\% & 78.4\%  & 85.8\% & 83.2\%  \\
  Ensemble & 5 CNNs &  71.7\% & 78.3\%  & 85.5\% & 83.2\%  \\
  \hline
  \end{tabular}
}
\end{center}
\vspace{-6pt}
\caption{\textbf{Full scene material classification results.} Mean class and
total accuracy on the test set.  When training, we optimize the CRF parameters
for mean class accuracy, but report both mean class and total accuracy (mean
accuracy across all examples).  In one experiment \textbf{(a)}, we train and test only on
segments; in a separate experiment \textbf{(b)}, we train and test only on
clicks.  Accuracies for segments are averaged across all pixels
that fall in that segment.
}
\label{tab:sceneaccuracy}
\vspace{-6pt}
\end{table}

We evaluate multiple versions of \googlenet: both the original architecture and
a version with the average pooling layer (at the end) changed to 5x5, 3x3, and
1x1 (i.e. no average pooling).  We evaluate \alexnet trained at multiple patch
scales (Figure~\ref{fig:patchsize}).  When using an \alexnet trained at a
different scale, we keep the same scale for testing.  We also experiment with
ensembles of \googlenet and \alexnet, combined with either arithmetic or
geometric mean.

Since we have two types of data, \emph{clicks} and \emph{segments}, we run two
sets of experiments: (a) we train and test only on segments, and in a separate
experiment (b) we train and test only on clicks.  These two training objectives
result in very different behavior, as illustrated in
Figure~\ref{fig:withoutcrf}.
In experiment (a), the accuracy across segments are optimized, producing clean
boundaries.  In experiment (b), the CRF maximizes accuracy only at click
locations, thus resulting in sloppy boundaries.  As shown in
Table~\ref{tab:sceneaccuracy}, the numerical scores for the two experiments are
also very different: segments are more challenging than clicks.
%
While clicks are sufficient to train a CNN, they are not sufficient to train a
CRF.

Focusing on segmentation accuracy, we see from Table~\ref{tab:sceneaccuracy}(a)
that our best single model is \googlenet without average pooling (6\% better than
with pooling).  The best ensemble is 2 CNNs: \googlenet (no average pooling) and
\alexnet (patch scale: 46.5\%), combined with arithmetic mean.  Larger ensembles
perform worse since we are averaging worse CNNs.  In
Figure~\ref{fig:sceneresults}, we show example labeling results on test images.


\subsection{Comparing \minc to FMD}

Compared to FMD, the size and diversity of \minc is valuable for
classifying real-world imagery. Table~\ref{tab:crossdataset} shows the
effect of training on all of FMD and testing on \minc (and vice
versa). The results suggests that training on FMD alone is not
sufficient for real-world classification.  Though it may seem that our
dataset is ``easy,'' since the best classifications scores are lower
for FMD than for \minc, we find that difficulty is in fact closely
tied to dataset size (Section~\ref{sec:datasetsize}).  Taking 100
random samples per category, \alexnet achieves 54.2 $\pm$ 0.7\% on
MINC (64.6 $\pm$ 1.3\% when considering only the 10 FMD categories)
and 66.5\% on FMD.

\subsection{Comparing CNNs with prior methods}
\label{text:fmdresult}
Cimpoi~\cite{cimpoi2014describing} is the best prior material classification
method on FMD. We find that by replacing DeCAF with oversampled \alexnet features we
can improve on their FMD results.  We then show that on \minc, a finetuned CNN
is even better.

To improve on~\cite{cimpoi2014describing}, we take their \textsc{SIFT\_IFV}, combine it with \alexnet fc7 features, and add oversampling~\cite{krizhevsky2012imagenet} (see supplemental for details). With a linear SVM we achieve 69.6 $\pm$ 0.3\% on FMD. Previous results are listed in Table~\ref{tab:fmdresults}.

Having found that \textsc{SIFT\_IFV}+fc7
is the new best on FMD, we compare it to a finetuned CNN on
a subset of \minc (2500 patches per category, one patch per photo).
Fine-tuning \alexnet achieves
76.0 $\pm$ 0.2\% whereas \textsc{SIFT\_IFV}+fc7 achieves
67.4 $\pm$ 0.5\% with a linear SVM (oversampling, 5 splits).
This
experiment shows that a finetuned CNN is a better method for \minc than \textsc{SIFT\_IFV}+fc7.


\begin{table}
\begin{center}
\tablesize{%
  \begin{tabular}{|ll|ll|}
    \hline
    & & \multicolumn{2}{c|}{\textbf{Test}} \\
    & & FMD & \minc \\
    \hline
    \multirow{2}{*}{\textbf{Train}}
         & FMD   & 66.5\% & 26.1\%  \\
         & \minc & 41.7\% & 85.0\%  \\
    \hline
  \end{tabular}
}
\tablesize{%
  \begin{tabular}{c}
    \textit{(10 categories} \\
    \textit{in common)}
  \end{tabular}
}
\end{center}
\vspace{-6pt}
\caption{%
  \textbf{Cross-dataset experiments.}  We train on one dataset and test on
  another dataset.  Since \minc contains 23 categories, we limit \minc to
  the 10 categories in common. CNN: \alexnet. 
}
\label{tab:crossdataset}
\vspace{-6pt}
\end{table}

\begin{table}
\begin{center}
\tablesize{%
  \begin{tabular}{|l|l|l|}
    \hline
    \textbf{Method} & \textbf{Accuracy} & \textbf{Trials} \\
    \hline
       Sharan \etal~\cite{sharan-IJCV2013} & 57.1 $\pm$ 0.6\% & 14 splits \\
       Cimpoi \etal~\cite{cimpoi2014describing} & 67.1 $\pm$ 0.4\% & 14 splits \\
    \hline
       Fine-tuned \alexnet & 66.5 $\pm$ 1.5\% & 5 folds \\
       \textsc{SIFT\_IFV}+fc7 & \textbf{69.6} $\pm$ \textbf{0.3\%} & 10 splits \\
    \hline
  \end{tabular}
}
\end{center}
\vspace{-6pt}
\caption{%
  \textbf{FMD experiments.} By replacing \decaf features with oversampled
  \alexnet features we improve on the best FMD result.
}
\label{tab:fmdresults}
\vspace{-6pt}
\end{table}

\section{Conclusion}
Material recognition is a long-standing, challenging problem. We
introduce a new large, open, material database, \OStwoshort, that
includes a diverse range of materials of everyday scenes and staged
designed interiors, and is at least an order of magnitude larger than
prior databases. Using this large database we conduct an
evaluation of recent deep learning algorithms for simultaneous
material classification and segmentation, and achieve results that surpass
prior attempts at material recognition.

Some lessons we have learned are:
\begin{packed_item}
\item Training on a dataset which includes the surrounding context is crucial
  for real-world material classification.
\item Labeled clicks are cheap and sufficient to train a CNN alone.  However, to
  obtain high quality segmentation results, training a CRF on polygons results
  in much better boundaries than training on clicks.
\end{packed_item}


Many future avenues of work remain. Expanding the dataset to a broader
range of categories will require new ways to mine images that have
more variety, and new annotation tasks that are cost-effective.
Inspired by attributes for textures~\cite{cimpoi2014describing}, in
the future we would like to identify material attributes and expand
our database to include them.  We also believe that further
exploration of joint material and object classification and
segmentation will be fruitful~\cite{hu2011toward} and lead to
improvements in both tasks. Our database, trained models, and all
experimental results are available online at \website.

\medskip
\noindent{\bf Acknowledgements.}
This work was supported in part by
Google,
Amazon AWS for Education,
a NSERC PGS-D scholarship,
the National Science Foundation (grants IIS-1149393, IIS-1011919, IIS-1161645),
and
the Intel Science and Technology Center for Visual Computing.

{\small
\bibliographystyle{ieee}
\bibliography{material}
}

\end{document}